\documentclass{article}

\PassOptionsToPackage{authoryear, round}{natbib}
\usepackage[preprint]{neurips_2026}      

\usepackage[utf8]{inputenc} 
\usepackage[T1]{fontenc}    
\usepackage{hyperref}       
\usepackage{url}            
\usepackage{booktabs}       
\usepackage{amsfonts}       
\usepackage{nicefrac}       
\usepackage{microtype}      
\usepackage{xcolor}         

\usepackage{amsmath}
\usepackage{bm}

\usepackage{amsmath,amsfonts,bm}

\def\eqref#1{equation~\ref{#1}}

\def\1{\bm{1}}

\DeclareMathAlphabet{\mathsfit}{\encodingdefault}{\sfdefault}{m}{sl}
\SetMathAlphabet{\mathsfit}{bold}{\encodingdefault}{\sfdefault}{bx}{n}

\usepackage{algorithm}
\usepackage{algpseudocode}

\usepackage{graphicx}
\usepackage{subcaption}

\usepackage{multirow}
\usepackage{threeparttable}
\usepackage{makecell}
\usepackage{array}
\usepackage{siunitx}      
\usepackage{enumitem}
\title{Dense2MoE: Pushing the Pareto Frontier of On-Device LLMs via Unified Pruning and Upcycling}

\author{
Fengfa Li$^{*1}$, 
Hongjin Ji$^{*1,2}$,
Yifeng Ding$^{1}$,
Lei Ren$^{1\dagger}$,
Chen Wei$^{1}$ \\
$^1$Li Auto \\
$^2$The Chinese University of Hong Kong, Shenzhen
}
\begin{document}

\maketitle
{\let\thefootnote\relax\footnotetext{$^*$Equal contribution.\quad $^\dagger$Corresponding author and project lead.}}

\begin{abstract}

The Mixture of Experts (MoE) architecture is highly promising for resource-constrained on-device deployments, yet training these models from scratch incurs prohibitive costs. Current methods attempt to alleviate this by upcycling dense models into MoEs; however, they often introduce parameter redundancy that degrades inference efficiency. Alternatively, standard layer pruning mitigates redundancy but inevitably compromises model accuracy. To resolve this dilemma, we propose \textbf{Dense2MoE}, a novel framework that unifies pruning and upcycling through \textbf{Layer-Fusion UpCycling} (LF-UC). Guided by hardware Roofline theory, Dense2MoE systematically overcomes the inference memory wall by pruning bandwidth-heavy attention modules from redundant layers, while repurposing their Multi-Layer Perceptrons (MLPs) into MoE experts. This structural innovation preserves the model’s core capabilities and strictly limits active parameters via selective token routing. With a modest continual pre-training budget, Dense2MoE efficiently converts publicly available dense LLMs into on-device-ready MoE models. Extensive experiments demonstrate that Dense2MoE significantly advances the Pareto frontier for on-device inference latency versus model accuracy, outperforming dense baselines, state-of-the-art compression, and standard upcycling methods.

\end{abstract}

\section{Introduction}
While large language models have achieved remarkable performance across diverse tasks~\citep{vaswani2017attention,touvron2023llama}, their deployment in latency-critical and resource-constrained environments---such as autonomous driving and edge AI---remains severely hindered. During the autoregressive decoding phase, dense LLMs are inherently memory-bound according to hardware Roofline theory~\citep{williams2009roofline}. Therefore, simply reducing theoretical FLOPs is insufficient to break the inference memory wall. Establishing a practical, industrial-standard recipe to repurpose these open-source dense LLMs into highly efficient architectures has become an urgent priority.

To alleviate deployment bottlenecks, the community has primarily explored two paradigms: model pruning and Mixture-of-Experts (MoE) upcycling. Structured layer pruning~\citep{gromov2024unreasonable} effectively curtails serial computation but inevitably sacrifices model accuracy by indiscriminately discarding multi-layer perceptrons (MLPs)—the primary knowledge repositories of LLMs~\citep{meng2022locating}.
Conversely, MoE upcycling~\citep{komatsuzaki2022sparse} transforms dense models into sparse architectures to scale capacity efficiently. However, early upcycling methods~\citep{komatsuzaki2022sparse} often suffer from homogeneous expert initialization, necessitating massive continual pre-training to induce specialization. 
Furthermore, recent sparse repurposing approaches~\citep{sukhbaatar2024branch, gao2025tomoe} predominantly target intra-layer redundancy. By neglecting substantial inter-layer structural overlaps, they fail to fully resolve the memory bandwidth bottlenecks imposed by redundant attention and MLP blocks.
 
To overcome this fundamental trade-off between inference efficiency and model accuracy, we propose Dense2MoE, a novel framework that seamlessly integrates structured layer pruning and MoE upcycling in a principled and unified manner. We first identify redundant decoder layers through a dual-constraint similarity metric that jointly evaluates layer-wise output similarity and MLP input feature matching. Rather than discarding pruned layers entirely, we physically eliminate their redundant attention modules to substantially reduce memory bandwidth consumption. Crucially, through our novel Layer-Fusion Upcycling (LF-UC) technique, we preserve the pre-trained MLPs from the pruned layers and integrate them as heterogeneous experts into the retained layers. This structural design ensures that the enhanced representational capacity introduces no additional dynamic latency overhead.  Furthermore, by initializing experts with heterogeneous pre-trained weights, Dense2MoE requires only a lightweight continual pre-training budget of 225B tokens, consuming less than 1\% of the compute required to train a MoE model from scratch.

In summary, our key contributions are as follows:
\begin{itemize}[leftmargin=*, nosep, topsep=0pt, partopsep=0pt]
    \item \textbf{A unified pruning and upcycling framework:} We propose Dense2MoE,  a unified framework that seamlessly integrates structured layer pruning with MoE upcycling via the LF-UC technique, enabling joint optimization of inference latency and model accuracy for resource constrained on-device LLM deployments.
    \item \textbf{State-of-the-art Pareto optimality:} Evaluated across 9 diverse benchmarks (e.g., MMLU, ARC, code generation), Dense2MoE advances the Pareto frontier of inference latency versus model accuracy. Grounded in hardware Roofline modelling, it outperforms dense baselines, state-of-the-art compression, and standard upcycling methods.
    \item \textbf{Robust scalability:} We validate the versatility of Dense2MoE across varying model scales (0.5B to 7B) and distinct architectural families (Qwen2.5 and LLaMA2), confirming its broad applicability for resource constrained on-device deployments.
\end{itemize}

\section{Related work}

\textbf{Structured Pruning for LLMs.}
Model pruning seeks to alleviate inference latency and memory constraints by systematically excising redundant weights. Recent efforts predominantly diverge into two trajectories: layer-wise pruning and fine-grained structural pruning. Layer-wise pruning methods, such as UIDL~\citep{gromov2024unreasonable}, attempt to drop consecutive redundant decoder layers based on block-similarity metrics. To mitigate the severe performance drop inherent to naive layer removal, subsequent methods like LLM-Streamline~\citep{chen2024streamlining} propose training lightweight surrogate modules (e.g., shallow FFNs) to compensate for the structural loss. Conversely, fine-grained approaches like LLM-Pruner~\citep{ma2023llm} selectively remove specific attention heads and FFN filters guided by gradient and activation heuristics.

Despite these advancements, existing pruning paradigms fundamentally struggle to preserve a model's advanced reasoning and coding capabilities. Specifically, gradient-based structural pruning often disrupts the pre-trained manifold, requiring intensive retraining, while layer-wise methods indiscriminately discard deep MLP modules alongside redundant attention mechanisms. Given that MLPs function as the primary knowledge repositories within Transformers~\citep{meng2022locating}, replacing them with shallow surrogate modules (as in LLM-Streamline) or discarding them entirely (as in UIDL) results in a permanent, unrecoverable loss of representational capacity. Dense2MoE diverges from this destructive paradigm: it physically prunes only the attention bottlenecks to strictly reduce memory bandwidth consumption, while strategically preserving and recycling the rich, pre-trained MLPs.

\textbf{MoE Upcycling and Architecture Repurposing.}
MoE upcycling~\citep{shazeer2017sparsely,komatsuzaki2022sparse} presents a compelling alternative to static pruning by expanding dense feed-forward networks into sparse, independent experts. This allows for increased model capacity without a commensurate linear increase in dynamic computational overhead. However, conventional upcycling methodologies—which typically initialize experts by directly duplicating native MLPs—suffer from severe representation collapse. Overcoming this homogeneous initialization necessitates massive continual pre-training to induce expert specialization, which directly contradicts the objective of low-cost model repurposing.

Furthermore, recent state-of-the-art sparse repurposing approaches, such as ToMoE~\citep{gao2025tomoe}, primarily focus on exploiting intra-layer dimensional redundancy within individual MLP modules. By neglecting the macro-level inter-layer structural redundancy inherent in deep Transformers, these approaches fail to address the predominant KV-cache loading overhead caused by redundant attention layers during autoregressive decoding. Dense2MoE overcomes these fundamental limitations by utilizing the preserved MLPs from structurally pruned layers as naturally heterogeneous experts. This Layer-Fusion Upcycling (LF-UC) strategy simultaneously eliminates attention-driven memory bottlenecks and bypasses the exorbitant training costs required for expert specialization.

\section{Method}
\begin{figure}[htbp]
    \centering
    \includegraphics[width=\textwidth]{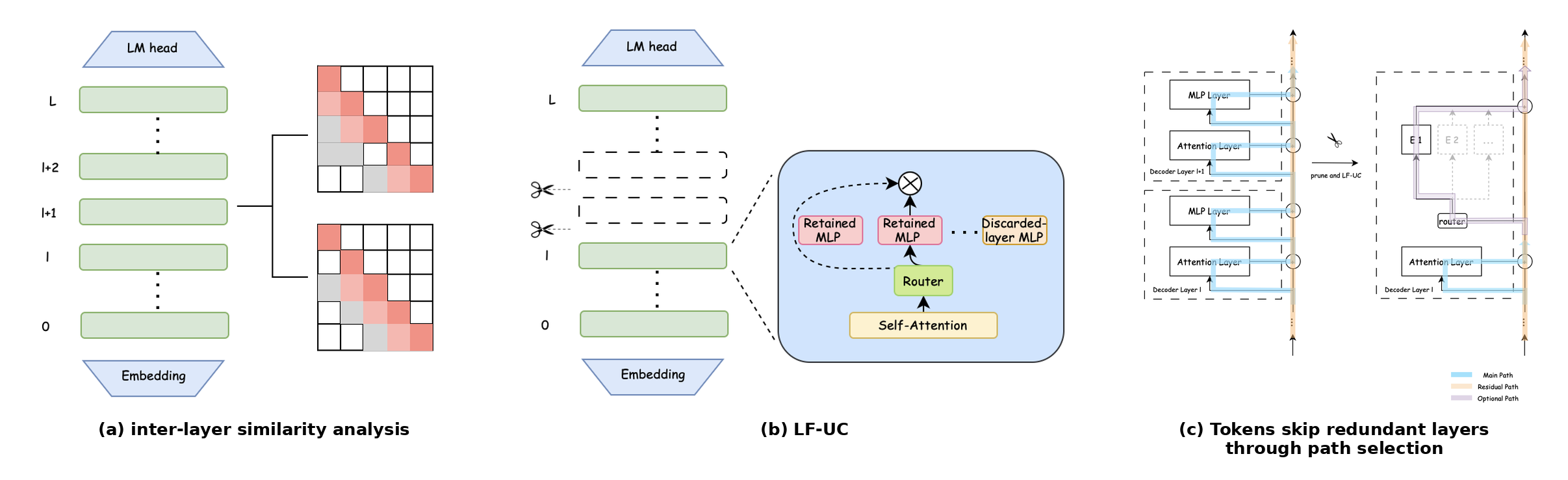}
    \caption{\textbf{Overview of Dense2MoE.} (a) Similarity analysis: Identifying structural redundancy via decoder output and MLP input similarities. (b) Layer-Fusion Upcycling (LF-UC): Pruning redundant attention modules while fusing their MLPs into retained blocks as MoE experts, preserving capacity with reduced compute. (c) Dynamic token-level routing: During inference, a router dispatches tokens to specific experts or allows them to skip the computational block via residual paths, effectively reducing active parameters and accelerating inference.}
    \label{fig1}
\end{figure}

Dense2MoE adopts a prune-and-expand approach to repurpose dense models into Mixture-of-Experts (MoE) architectures, aiming to reduce active parameters and accelerate inference without compromising performance. As illustrated in Figure \ref{fig1}, our method can be summarized into three key stages:

(1) Inter-layer Similarity Analysis ((a)): We identify structural redundancy by localizing adjacent layers that exhibit high cosine similarity in both decoder outputs and MLP inputs.\\
(2) LF-UC ((b) and (c)): For a set of identified redundant layers, we retain the first block as the base. We completely prune the attention modules of subsequent layers to reduce compute, and aggregate their MLPs into the base block as distinct experts. Simultaneously, a dynamic token-level router is introduced to selectively dispatch tokens to these upcycled experts or bypass them entirely via residual paths, structurally enabling faster inference.\\
(3) Continual Pre-Training: Following the architectural modifications, we conduct a continual pre-training phase to optimize the routing mechanism, align the newly fused experts, and fully recover the model's capabilities.

Throughout the subsequent formulations, layer indices are consistently denoted by superscripts and token dimensions by subscripts (a comprehensive notation guide is provided in Table~\ref{tab:core_notation_detailed} in Appendix~\ref{app:notation}).

\subsection{Preliminary}

\textbf{Forward Propagation in LLMs.}
Modern decoder-only large language models (LLMs) primarily stack transformer blocks along the depth dimension. The inter-layer information transmission relies on residual connections. For the $l$-th layer, the hidden state $x^{(l)}$ is updated via Multi-Head Attention (MHA) and Multi-Layer Perceptron (MLP) modules:
\begin{equation}
    x^{(l+1)} = h^{(l)} + \text{MLP}^{(l)}(\text{LN}(h^{(l)}), \theta_{mlp}^{(l)})
\end{equation}
where $h^{(l)} = x^{(l)} + \text{MHA}^{(l)}(\text{LN}(x^{(l)}), \theta_{mha}^{(l)})$ is the intermediate state post-attention, and $\text{LN}(\cdot)$ denotes standard Layer Normalization.

Crucially, the MLP module, which typically adopts a Gated Linear Unit (GLU) variant in mainstream LLMs, serves as the core repository of learned knowledge reused in our method. The transformation is defined as:
\begin{equation}
    \label{eq:mlp}
    \text{MLP}^{(l)}(h, \theta_{mlp}^{(l)}) = \left( \sigma \left( h W_{up}^{(l)} \right) \odot \left( h W_{gate}^{(l)} \right) \right) W_{down}^{(l)}
\end{equation}
where $W_{up}^{(l)}, W_{gate}^{(l)} \in \mathbb{R}^{d \times d_{mid}}$ are the up-projection and gate matrices, $W_{down}^{(l)} \in \mathbb{R}^{d_{mid} \times d}$ is the down-projection matrix, and $\sigma(\cdot)$ denotes the SiLU activation. Since the layer's representational capacity is entirely encoded within these three weight matrices, our approach directly preserves and reallocates them from redundant layers to construct expert modules, avoiding any disruptive structural modifications.

\textbf{Mixture-of-Experts (MoE).} 
In sparse MoE architectures, the standard dense MLP block is replaced by a routing network and a set of $N$ independent experts $\{E_1, \dots, E_N\}$. For an input token representation $h$, a trainable router computes a probability distribution $g(h) = \text{Softmax}(h W_r)$, parameterized by routing weights $W_r \in \mathbb{R}^{d \times N}$. To ensure inference efficiency, conditional computation is enforced by selecting only the Top-$k$ experts per token.The MoE output is thus the sparsity-weighted sum of these activated experts:
\begin{equation}
    \text{MoE}^{(l)}(h) = \sum_{i \in \text{Top-}k} g_i(h) \cdot E_i^{(l)}(h)
\end{equation}
Crucially, each expert $E_i^{(l)}(\cdot)$ shares the identical structural parameterization as the dense MLP defined in Eq.~(\ref{eq:mlp}).To ensure uniform feature distributions for the router and strictly constrain the parameter count, all upcycled experts within the $l$-th layer share a single, unified Layer Normalization module.This architectural equivalence is the foundational premise enabling Dense2MoE to directly repurpose redundant dense layers into sparsely activated MoE experts.

\subsection{Inter-Layer Similarity Analysis and Hardware-Aware Trade-off}
\label{sec:similarity_and_tradeoff}

Analyzing inter-layer similarity is crucial for our pruning-and-expansion framework. Prior approaches typically rely on cosine similarity, which captures directional alignment but is strictly scale-invariant. To ensure activation stability during expert routing, we comprehensively evaluate both directional similarity and absolute scale discrepancy.

For a base layer $l$ and a subsequent layer $l+n$, we compute the sequence-averaged cosine similarities for both the final layer outputs ($y_t$) and the MLP inputs ($h_t$) across $T$ tokens:
\begin{equation}
    s_{out} = \frac{1}{T} \sum_{t=1}^T s_{cos}\left(y_t^{(l)}, y_t^{(l+n)}\right), \quad s_{mlp} = \frac{1}{T} \sum_{t=1}^T s_{cos}\left(h_t^{(l)}, h_t^{(l+n)}\right).
\end{equation}

A layer $l+n$ is identified as structurally redundant if it satisfies the directional constraints $s_{out} > 1 - \delta$ and $s_{mlp} > 1 - \delta$, where $\delta$ is a strict similarity threshold. Unlike prior methods that discard redundant blocks entirely \citep{gromov2024unreasonable,yang2024laco}, we selectively share attention layers with high $s_{mlp}$ (Section \ref{s23}) and retain their MLPs to preserve critical model capacity. 

However, directly routing to these repurposed MLPs risks activation anomalies if their expected input magnitudes differ significantly. Therefore, we impose a strict feature norm matching constraint:
\begin{equation} \label{eq:norm_constraint}
    \Delta_{norm} \left(h^{(l)}, h^{(l+n)}\right) = \frac{\big| \|h^{(l)}\|_2 - \|h^{(l+n)}\|_2 \big|}{\|h^{(l+n)}\|_2} < \epsilon.
\end{equation}
This mathematically guarantees that reusing redundant MLPs introduces no extreme scale shifts.As empirically verified in Appendix~\ref{app:ablation_norm}, omitting this constraint destabilizes the initial training phase, confirming its necessity for numerical consistency.We visualize the distribution of $s_{out}$ and $s_{mlp}$ across all layers for representative models in Fig.~\ref{fig3} (Appendix~\ref{app:similarity_viz}), where grid cell colors denote hidden state distances, establishing a strong empirical prior for structural redundancy.

\textbf{Mapping Thresholds to Topological Depth.} 
Based on these strictly defined constraints, we formulate a greedy iterative search algorithm (Algorithm~\ref{alg:global_search} in Appendix~\ref{app:algorithm}) to partition the network into a retained base layer set ($\mathcal{L}_{keep}$) and a pruned redundant layer set ($\mathcal{L}_{prune}$). Crucially, rather than treating the hyperparameters as isolated variables, this algorithm establishes a discrete mapping function between the constraints and the ultimate pruning depth: $m = f(\delta, \epsilon)$. Relaxing the cosine similarity threshold ($\delta$) and the norm error tolerance ($\epsilon$) acts as a continuous compression valve. This monotonically increases the topological depth of pruning, resulting in a step-wise phase transition in the model's physical structure (detailed sensitivity analysis provided in Appendix~\ref{app:threshold_sensitivity}). This mapping yields a set of candidate configurations on the Pareto frontier, necessitating a principled mechanism for optimal depth selection.

\textbf{Hardware-Aware Capacity-Efficiency Trade-off.} 
To systematically determine the optimal architectural ``sweet spot'' from these candidate depths, we adopt a hardware-aware trade-off metric inspired by MnasNet~\citep{tan2019mnasnet}:
\begin{equation} \label{eq:reward}
    \text{Reward}_i = S_i \times \left( \frac{LaT_i}{LaT_{base}} \right)^{w}
\end{equation}
where $S_i$ and $LaT_i$ are the candidate's average score and physical latency, and $LaT_{base}$ serves as the dense baseline's latency anchor. The exponent $w$ is an empirically determined constant that controls the strictness of the latency penalty. 

Following the methodological principle of MnasNet, we establish this hyperparameter based on the empirical scaling rule of the specific deployment hardware. An empirical rule for picking $w$ is to ensure that Pareto-optimal solutions yield similar rewards under different accuracy-latency trade-offs. This mathematically derived exponent enforces a rigorous balance between inference acceleration and capacity degradation. By evaluating all candidate configurations against this metric, we systematically identify and lock the optimal retained depth that achieves the maximum overall reward for subsequent MoE upcycling.

    

\subsection{Layer Fusion and Upcycling (LF-UC)}
\label{s23}

For a retained base layer $l^*$ and its corresponding $n^*$ redundant layers, the similarity constraint $h^{(l^*+i)} \approx h^{(l^*)}$ (where $i \in \{1, \dots, n^*\}$) dictates that the sequential Multi-Head Attention (MHA) computations across these blocks are highly repetitive. 

To exploit this, we mathematically fuse these $n^*+1$ sequential layers into a single MoE layer. We construct an expert pool of $N = K + n^* \times M$ experts by duplicating the native MLP of the base layer $K$ times (base experts) and repurposing the MLPs of the redundant layers $M$ times (supplementary experts). Distinct from prior upcycling methods that partition weights within a single layer, our approach leverages inter-layer structural heterogeneity, treating entire redundant MLPs as distinct, specialized experts.
This inter-layer heterogeneity also mitigates the Winner-Takes-All (WTA) routing collapse---wherein a single expert monopolises token dispatch while others remain underutilised---that afflicts homogeneous upcycling baselines (see Appendix~\ref{app:wta}).

During the forward pass, we bypass the $n^*$ redundant MHA computations, utilizing $h^{(l^*)}$ as the shared routing input:
\begin{equation}
    y^{(l^*+n^*)} = h^{(l^*)} + \sum_{j \in \mathcal{K}(h^{(l^*)})} g_j\left(h^{(l^*)}\right) \cdot \mathrm{MLP}_j\left(\mathrm{LN}_{mlp}\left(h^{(l^*)}\right)\right)
\end{equation}
By sharing the base MHA, this structural fusion formally eliminates $n^*$ redundant attention operations—significantly mitigating dynamic computational bottlenecks—while fully preserving the pre-trained knowledge embedded across all original MLPs.

\section{Experiments}

We validated the Dense2MoE method on models of different scales across two open-source model families (Qwen2.5~\citep{qwen25} and Llama2~\citep{touvron2023llama}), and compared it with prevalent model layer pruning and sparse repurposing approaches. For fairness, we constructed a 225B-token dataset with the same composition to compare model performance. All methods were fine-tuned on the dataset, and their performance was validated on benchmarks spanning multiple domains (mathematics, code, reasoning and general knowledge). 

\subsection{Setup}
\label{sec:setup}
To validate our approach at scales relevant to edge deployment, we conduct experiments on Qwen2.5-0.5B~\citep{qwen25}. To assess generality, we further extend our method to larger and heterogeneous model families, including Qwen2.5-1.5B~\citep{qwen25} and Llama2-7B~\citep{touvron2023llama}. Unless otherwise specified, we use the following setup: all models are trained via continual pre-training on 225B tokens with a batch size of 40M tokens. We adopt a linear warm-up from 0 to $1\times10^{-4}$ over the first 360M tokens, followed by cosine decay to $1\times10^{-5}$.

For MoE-specific optimization, we use an auxiliary load-balancing loss~\citep{shazeer2017outrageously, fedus2022switch} weighted by $1\times10^{-3}$. We configure the MoE layer with $N=6$ experts and top-$k$ routing with $k=1$~\citep{fedus2022switch}. We further ablate the effect of varying the number of experts $N$ in Section~\ref{sec:ablation_experts}.

\textbf{Evaluation benchmarks.} We verified the effectiveness of our method on general knowledge benchmarks including C-Eval~\citep{huang2023c}, CMMLU~\citep{li2023cmmlu}, and MMLU~\citep{hendrycks2020measuring}. To evaluate mathematical and coding capabilities, we utilized GSM8K~\citep{liu2023tinygsm}, CMath~\citep{wei2023cmath}, HumanEval~\citep{chen2021evaluating}(abbreviated as HEval in tables), and MBPP~\citep{austin2021program}. For assessing complex reasoning, we adopted BBH~\citep{suzgun2022challenging} and ARC-Challenge~\citep{clark2018think}.

\textbf{Baselines.} To comprehensively evaluate Dense2MoE, we benchmark against the original dense seed models and several representative state-of-the-art compression paradigms. These include layer-pruning methods (\textbf{UIDL}~\citep{gromov2024unreasonable};\textbf{LLM-Streamline}~\citep{chen2024streamlining}), fine-grained structural pruning (\textbf{LLM-Pruner}~\citep{ma2023llm}), and conventional \textbf{MoE Upcycling}~\citep{komatsuzaki2022sparse}. Furthermore, we compare against recent advanced sparse repurposing frameworks, specifically \textbf{LLaMA-MoE}~\citep{zhu2024llama} and \textbf{ToMoE}~\citep{gao2025tomoe}, which focus on partitioning feed-forward networks and exploiting intra-layer dimensional redundancy, respectively.

\textbf{Hardware and inference evaluation setup.} Considering the target application scenarios of edge-side and embodied AI models, we evaluate the inference efficiency targeting the NVIDIA Jetson Thor-U automotive SoC (Peak FP16 compute $\pi_H = 350$~TFLOPS, memory bandwidth $\beta_H = 273$~GB/s). In practical edge deployment, empirical latency measurements are frequently distorted by suboptimal software-level implementations, such as unoptimized MoE routing kernels in current inference engines, which obscure the true architectural efficiency. To conduct a rigorous and hardware-agnostic evaluation, we calculate the theoretical end-to-end physical latency bounds based on the first-principles Roofline model~\citep{sun2026hardware}. 

Our evaluation workload ($B=1$, $S_{in}=1000$, $S_{out}=50$) mirrors realistic automotive and embodied AI scenarios, which typically process extensive context to generate concise control commands. Crucially, this end-to-end setting captures Dense2MoE's dual acceleration: pruning redundant attention significantly reduces FLOPs in the compute-bound \textit{prefill} phase, while sparse expert routing bypasses the memory wall in the memory-bound \textit{decode} phase (detailed Roofline analysis in Appendix~\ref{app:roofline}).

\subsection{Dataset details}
Our dataset comprises approximately 225B tokens from publicly available, open-source sources, which we meticulously cleaned, normalized, and deduplicated. To better align the Dense2MoE-modified architecture with diverse downstream tasks, we carefully adjusted domain proportions for targeted re-adaptation. Detailed composition and sampling ratios are provided in Table~\ref{tab:data-mixture-app} (see Appendix~\ref{app:data}).

\subsection{Main Results}

\begin{figure}[htbp]
    \centering
    \includegraphics[height=0.25\textheight, keepaspectratio]{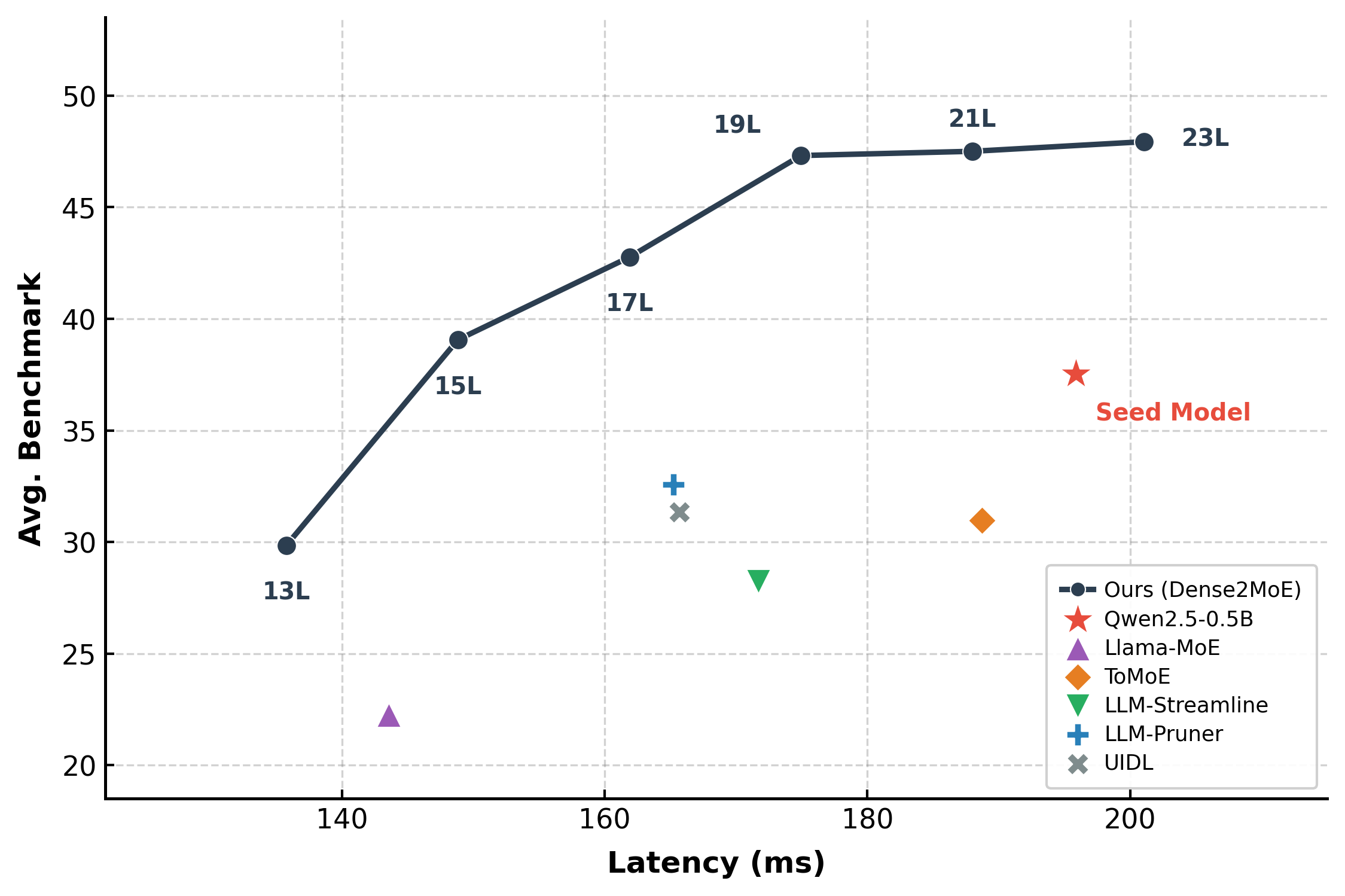}
    \caption{Pareto frontier of benchmark performance versus physical inference latency on the Thor-U platform.}
    \label{pareto}
\end{figure}

As illustrated in Figure~\ref{pareto}, sweeping the layer depth under these constraints reveals that retaining 19 layers yields the optimal balance. Compared with the dense baseline and existing alternatives, this optimal Dense2MoE configuration ($L=19$) significantly advances the Pareto frontier of benchmark performance versus inference latency.The data points generated by our approach strictly dominate the trade-off space. Most notably, Dense2MoE comprehensively surpasses the original 24-layer Seed Model (Qwen2.5-0.5B, latency:195.87ms, average score:37.53), simultaneously achieving higher benchmark accuracy and substantially lower physical inference time.

Beyond theory, empirical test results on the Thor-U SoC ($B=1$, $S_{in}=1000$, $S_{out}=50$) confirm that our optimal Dense2MoE ($L=19$) achieves a total latency of 271.0\,ms, dropping significantly from the seed model's 307.5\,ms. This speedup, achieved without optimized sparse MoE kernels, highlights immense potential for future hardware-level acceleration.

To identify the absolute optimal configuration from this frontier, we instantiate the hardware-aware reward metric Eq.(~\ref{eq:reward}) on the target Thor-U platform. On this specific hardware, we empirically observe that doubling the inference latency typically brings about an $11\%$ relative benchmark gain. To satisfy the constant reward constraint ($2^{-w} = 1.11$), we establish the platform-specific penalty exponent as $w \approx -0.15$. Evaluated against this stringent metric, the 19-layer architecture achieves the maximum overall reward (detailed quantitative evaluations are provided in Appendix~\ref{app:reward_details}). Consequently, we lock this depth ($L=19$) as the optimal foundation for our subsequent MoE expansions.

Furthermore, when evaluated against recent state-of-the-art (SOTA) pruning and MoE upcycling methods, Dense2MoE consistently establishes a superior efficiency envelope. While traditional pruning and existing sparse repurposing approaches (such as ToMoE and Llama-MoE) successfully reduce physical inference latency, they suffer from severe capacity collapse,particularly catastrophic degradation in complex reasoning and coding tasks. As shown in Table~\ref{tab:performance_comparison}, these baselines fail to preserve the seed model's reasoning capabilities, often falling significantly below the original dense performance. In contrast, the Dense2MoE Pareto frontier demonstrates robust representational retention. This systematic improvement confirms that our Layer-Fusion Upcycling effectively eliminates redundant memory-bound attention operations without destroying pre-trained knowledge.Consequently, Dense2MoE achieves faster inference by minimizing active parameters (0.42B) and substantially surpasses the original dense model's average score, all while remaining strictly within edge memory capacity limits (total parameters $\approx 1.66$B).This provides a highly competitive and flexible continuum of models optimal for latency-critical edge deployment.

\begin{table}[htbp]
    \centering
    \scriptsize
    \setlength{\tabcolsep}{2pt}
    \renewcommand{\arraystretch}{0.95}
    \caption{Performance--efficiency trade-off across benchmarks.}
    \label{tab:performance_comparison}
    \begin{tabular}{l c c | *{9}{c} | c}
        \toprule
        \multirow{2}{*}{Method} 
        & \multirow{2}{*}{Act.Param.} 
        & \multirow{2}{*}{Latency(ms)} 
        & \multicolumn{9}{c|}{Benchmarks} 
        & \multirow{2}{*}{Avg.} \\
        \cmidrule(lr){4-12}
        & & & CEVal & CMMLU & MMLU & CMath & GSM8K & HEval & MBPP & BBH & ARC-C \\
        \midrule
        
        \multicolumn{13}{c}{Dense / Pruned Models} \\
        \midrule
        Qwen2.5-0.5B 
        & 0.50B & 195.87 
         & 53.77 & 52.06 & 47.60 & 30.66 & 38.59 & 28.66 & 29.60 & 29.85 & 26.96
        & 37.53 \\
        
        UIDL 
        & 0.42B & 165.68 
        & 43.63 & 45.28 & 36.10 & 30.50 & 26.54 & 27.44 & 21.80 & 27.50 & 23.29 
        & 31.34 \\
        
        LLM-Pruner 
        & 0.42B & 165.22 
        & 50.01 & 46.20 & 40.70 & 30.00 & 34.19 & 10.37 & 32.80 & 22.16 & 26.79 
        & 32.58 \\
        
        LLM-Streamline 
        & 0.44B & 171.72 
        & 46.00 & 43.09 & 39.10 & 27.33 & 23.88 & 7.90 & 26.40 & 17.64 & 22.78 
        & 26.79 \\
        
        \midrule
        
        \multicolumn{13}{c}{Sparse MoE Models} \\
        \midrule
        ToMoE 
        & 0.43B & 188.7 
        & 49.50 & 48.05 & 40.30 & 32.33 & 21.53 & 12.80 & 19.40 & 28.12 & 26.79 
        & 30.98 \\
        
        Llama-MoE 
        & 0.34B & 143.56 
        & 42.26 & 41.15 & 37.40 & 25.67 & 6.82 & 6.70 & 6.40 & 11.14 & 22.70 
        & 22.25 \\
        
        Dense2MoE 
        & 0.42B & 174.95 
        & 62.68 & 64.10 & 45.00 & 45.03 & 44.67
        & 42.68 & 43.00 & 36.63 & 42.03
        & \textbf{47.31} \\
        
        \bottomrule
    \end{tabular}
    \vspace{0.2em}
    \begin{minipage}{0.95\linewidth}
        \centering
        \scriptsize 
        \textbf{Note:} Activated Parameters denote per-token inference cost. 
        MoE models have higher total parameters but maintain comparable activated cost via sparse routing.
    \end{minipage}
\end{table}

Building upon the Pareto dominance established above, we further dissect the cognitive capabilities of our optimal Dense2MoE configuration against other methods. While the Pareto analysis demonstrates macro-level efficiency, Table~\ref{tab:performance_comparison} provides a granular performance breakdown across nine diverse benchmarks encompassing general knowledge, mathematics, coding, and reasoning.

As the detailed results indicate, the superiority of Dense2MoE is consistent across distinct cognitive domains rather than being skewed by a single metric. Notably, after continual pre-training on 225B tokens, the optimal Dense2MoE configuration consistently outperforms competing methods. While activating only 80\% of the seed model's parameters, it yields an absolute average improvement of over 9.78 percentage points compared to the dense baseline. This comprehensive evaluation robustly validates that our structural pruning and layer-fusion mechanism perfectly preserves, and even enhances, the model's multidimensional reasoning capacity under strict inference constraints.

\textbf{Generalization and scalability analysis.} To validate the generality and scalability of our method, we extend Dense2MoE to models of different families and sizes, including LLaMA2-7B and Qwen2.5-1.5B. The results are presented in Table~\ref{tab:generalization_scaling}. After fine-tuning Dense2MoE using only 225B tokens, it outperforms Qwen2.5-1.5B by 3.04 percentage points and LLaMA2-7B by 3.54 percentage points, demonstrating that our model remains effective as model scale increases and can be extended to other open-source model families.

\begin{table}[htbp]
    \centering
    \footnotesize 
    \setlength{\tabcolsep}{2.5pt}
    \caption{Performance comparison of different methods across various benchmarks.}
    \label{tab:generalization_scaling}
    \begin{tabular}{l l c *{9}{c} c}
        \toprule
        \multirow{2}{*}{LLM} & \multirow{2}{*}{Model} & \multirow{2}{*}{\makecell{Activated\\Parameters}} & \multicolumn{9}{c}{Benchmarks} & \multirow{2}{*}{Avg.} \\
        \cmidrule(lr){4-12}
        & & & CEVAL & CMMLU & MMLU & CMath & GSM8K & HEval & MBPP & BBH & ARC-C \\
        \midrule
        \multirow{2}{*}{Qwen2.5}
        & Dense & 1.5~B & 68.72 & 67.82 & 61.10 & 63.99 & 49.00 & 35.98 & 46.00 & 39.73 & 64.42 & 55.19\\
        & Ours& 1.2~B & 72.51 & 70.5 & 51.67 & 55.5 & 63.31 & 52.44 & 53.2 & 42.73 & 62.2 &\textbf{58.23}\\
        \midrule
        \multirow{2}{*}{Llama2} 
        & Dense & 7.0~B & 30.99 & 32.75 & 45.80 & 16.60 & 22.83 & 12.80 & 21.60 & 39.36 & 27.80 & 27.84 \\
        & Ours & 5.7~B & 38.68 & 39.56 & 50.8 & 25.33 & 26.31 & 18.29 & 22.4 & 34.59 & 26.45 & \textbf{31.38} \\
        \bottomrule
    \end{tabular}
\end{table}

\section{Analysis}
\subsection{Effectiveness of Layer-Fusion Upcycling}
\label{sec:lfuc_effectiveness}
\textbf{LF-UC achieves the best performance under the edge-latency constraint.} As shown in Table~\ref{tab:ablation}, Non-Prune+UC scores marginally higher (48.19 vs.\ 47.31), but its 24-layer depth incurs 207.58~ms latency---19\% above the deployment target---rendering it infeasible for edge deployment. At iso-latency (174.95~ms), LF-UC outperforms Prune+UC by 6.56 points, demonstrating that the advantage stems from heterogeneous expert initialisation rather than increased model depth.

LF-UC also leads to more balanced expert routing. As shown in Appendix~\ref{app:wta}, it reduces the mean dominant-expert load from 48.0\% to 45.2\% and decreases the number of layers exceeding the 50\% threshold from 11 to 9 compared to homogeneous Prune+Upcycling. We attribute this improvement to the source diversity of experts, where initialization from different layers introduces natural specialization and mitigates early-stage routing collapse.

\begin{table}[htbp]
    \centering
    \caption{Ablation on upcycling strategies (UC = Upcycling).}
    \label{tab:ablation}
    \scriptsize
    \setlength{\tabcolsep}{2pt}
    \renewcommand{\arraystretch}{0.95}
    \begin{tabular}{l c c | *{9}{c} | c}
        \toprule
        \multirow{2}{*}{Method}
          & \multirow{2}{*}{Act.Param.}
          & \multirow{2}{*}{Latency(ms)}
          & \multicolumn{9}{c|}{Benchmarks} & \multirow{2}{*}{Avg.} \\
        \cmidrule(lr){4-12}
        & & & C-Eval & CMMLU & MMLU & CMath & GSM8K & HEval & MBPP & BBH & ARC-C \\
        \midrule
        Non-Prune+UC & 0.49B & 207.58 & 62.02 & 60.81 & 45.6 & 47.0 & 51.2  & 40.73 & 44.6 & 39.0  & 42.78 & \textbf{48.19} \\
        Prune+UC     & 0.42B & 174.95 & 59.63 & 57.45 & 43.5 & 43.5 & 46.55 & 19.51 & 41.2 & 25.89 & 29.52 & 40.75 \\
        LF-UC (ours) & 0.42B & \textbf{174.95} & 62.68 & 64.1  & 45.0 & 45.03 & 44.67 & 42.68 & 43.0 & 36.63 & 42.03 & 47.31 \\
        \bottomrule
    \end{tabular}
\end{table}

\subsection{Ablation on Expert Scaling}
\label{sec:ablation_experts}

\begin{figure}[htbp]
    \centering
    \includegraphics[width=0.62\linewidth]{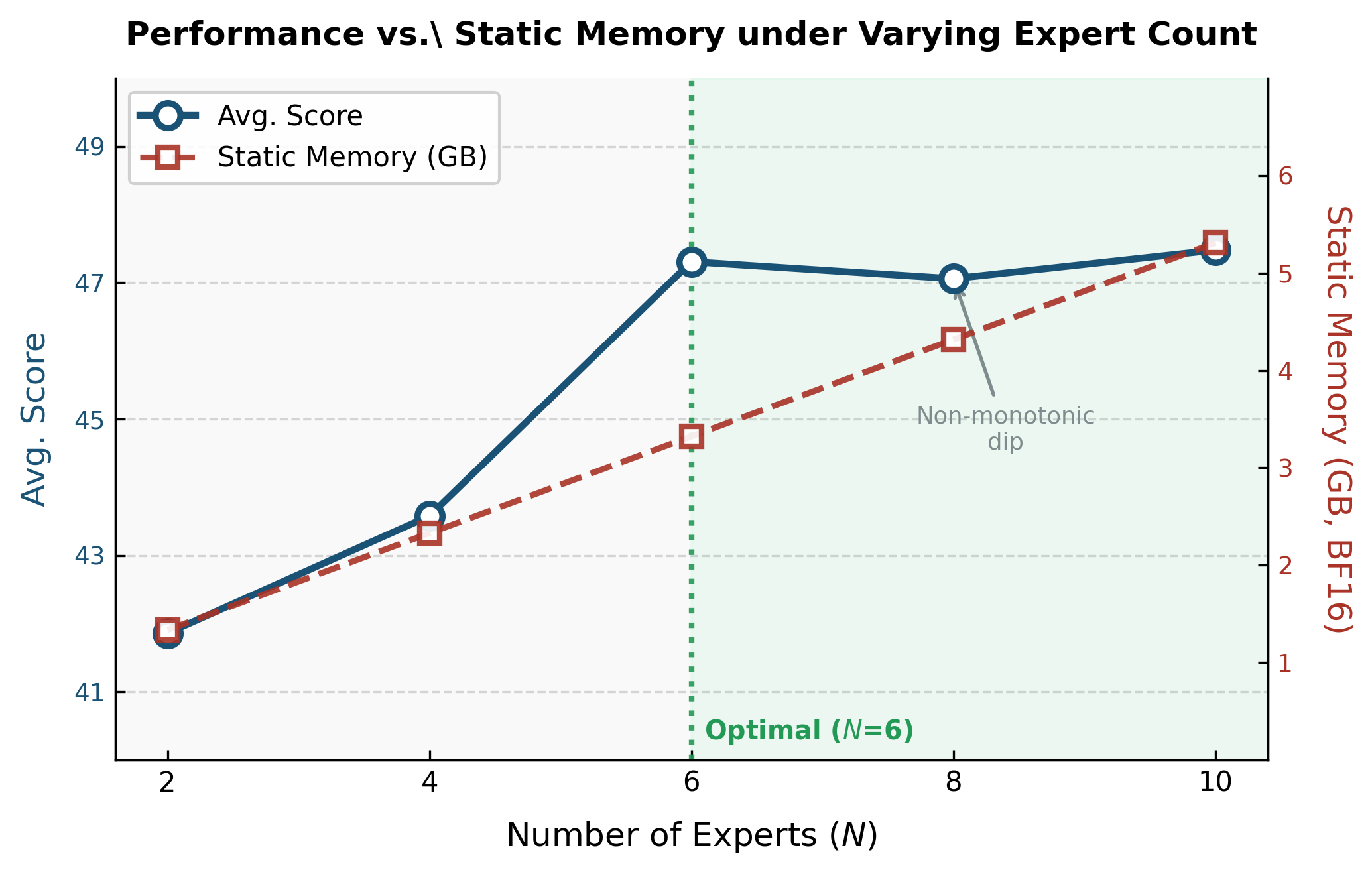}
    \caption{Effect of expert count ($N$) on average score (left axis, blue) and static parameter memory in BF16 (right axis, red), at fixed depth $L=19$.}
    \label{fig:expert_scaling}
\end{figure}

Fixing the optimal retained depth at $L=19$, we isolate the impact of scaling the total number of experts ($N$) under top-1 routing. As shown in Figure~\ref{fig:expert_scaling}, performance improves sharply from $N=2$ to $N=6$ but saturates and fluctuates beyond that point. $N=10$ nominally achieves the highest score, yet the trajectory is non-monotonic (a dip occurs at $N=8$) and the net gain over $N=6$ is less than 0.2 points.

Crucially, while inactive experts incur zero runtime memory-bandwidth cost, each additional expert linearly inflates the static parameter memory (right axis, Figure~\ref{fig:expert_scaling}; derivation in Appendix~\ref{app:memory_formula}). Scaling from $N=6$ to $N=10$ expands the model from 3.32~GB to 5.31~GB (BF16), a 60\% increase with negligible accuracy return. On automotive edge SoCs (e.g., Jetson Thor-U) with Unified Memory Architecture, this additional 2~GB must be carved from a shared LPDDR budget also used by sensor fusion and motion-planning pipelines, risking system-level memory pressure. We therefore adopt $N=6$ as the hardware-aware default for Dense2MoE.

\subsection{Efficiency-Capacity Trade-off in Edge Deployment}
\label{sec:efficiency_tradeoff}

To validate robustness under strict edge-latency budgets ($\sim$175~ms), we analyze the training dynamics of the extreme heterogeneous configuration \textbf{n1k4m2}---where $n^*=1$ redundant layer is fused, utilizing $K=4$ base experts and $M=2$ supplementary experts. Despite its aggressive compression, \textbf{n1k4m2} seamlessly avoids capacity bottlenecks and routing collapse, demonstrating two critical properties: 

(1) \textbf{Zero capacity penalty:} It mirrors the convergence trajectory of computationally heavier baselines (e.g., \textbf{n2k2m2}), achieving a final Language Modeling Loss of 1.2025.\\ 
(2) \textbf{Routing stability:} The Load Balancing Loss rapidly stabilizes within a healthy margin (2.0--2.5), explicitly preventing expert starvation. 

Consequently, these microscopic training dynamics confirm that \textbf{n1k4m2} maximizes representational power strictly within Roofline latency boundaries (detailed trajectories provided in Appendix~\ref{app:training_dynamics}).

\section{Conclusion}
In this work, we propose Dense2MoE, a unified framework that fundamentally resolves the accuracy–latency trade-off for resource-constrained on-device LLMs. Through our Layer-Fusion Upcycling (LF-UC) technique, Dense2MoE physically eliminates redundant attention layers to substantially reduce memory-bound inference latency, while strategically repurposing pruned MLPs as heterogeneous experts to preserve representational capacity. By systematically exploiting inter-layer structural redundancy, Dense2MoE enables joint optimization of inference latency and model accuracy, advancing the Pareto frontier beyond dense baselines, state-of-the-art compression methods, and standard upcycling approaches across 9 diverse benchmarks. Validated across varying model scales (0.5B to 7B) and diverse architectural families (Qwen2.5 and LLaMA2), Dense2MoE establishes a practical, hardware-aware paradigm for Pareto-optimal LLM deployment in resource-constrained edge AI systems.







\newpage
\bibliographystyle{plainnat}
\bibliography{neurips_2026_reference}     

\newpage 

\appendix
\section{Core Notation and Definitions}
\label{app:notation}

\begin{table}[htbp]
\centering
\caption{Core notation and strict definitions}
\label{tab:core_notation_detailed}
\renewcommand{\arraystretch}{1.2}
\begin{tabular}{lp{7cm}p{5.5cm}}
\toprule
\textbf{Symbol} & \textbf{Strict Definition} & \textbf{Dimension / Default Value} \\
\midrule
$L$ & Total number of Transformer decoder layers in the native dense LLM & Positive integer, e.g., 24 for Qwen2.5-0.5B \\
$l$ & Decoder layer index & $l \in \{1,2,\dots,L\}$ \\
$T$ & Sequence length of input text (number of tokens) & Positive integer, default 2048 \\
$d$ & Transformer hidden dimension (hidden size) & Positive integer, e.g., 896 for Qwen2.5-0.5B \\
$d_{mid}$ & MLP intermediate dimension (feed-forward dim) & Positive integer, e.g., 4864 for Qwen2.5-0.5B \\
$x^{(l)}$ & Input hidden state of the $l$-th decoder layer & Shape $T \times d$, $x^{(1)}$ is word embedding + positional encoding input \\
$h^{(l)}$ & Input hidden state of the $l$-th MLP layer (Attention output + residual connection) & Shape $T \times d$, core shared variable \\
$y^{(l)}$ & Final output of the $l$-th decoder layer & Shape $T \times d$, satisfying $y^{(l)} = x^{(l+1)}$ \\
$\theta_{mha}^{(l)}$ & All weights of the $l$-th layer Multi-Head Attention (MHA) & Including $W_Q, W_K, W_V, W_O$ and LN parameters \\
$\theta_{mlp}^{(l)}$ & All weights of the $l$-th layer MLP & Including gate, up-projection, down-projection matrices and LN parameters \\
$W_r$ & Routing weight matrix for the MoE layer & Shape $d \times N$ \\
$g_i(h)$ & Routing probability (gating score) assigned to the $i$-th expert for token $h$ & Real value in $(0, 1)$, $\sum g_i = 1$ \\
$E_i^{(l)}$ & The $i$-th independent expert module in the $l$-th layer & Identical parameterization to $\theta_{mlp}^{(l)}$ \\
$l^*$ & Base index of the selected retained layer & Positive integer, $1 \le l^* < L$ \\
$n^*$ & Total number of prunable redundant layers corresponding to a single retained layer & Positive integer, $l^*+n^* \le L$ \\
$N$ & Total number of experts in the fused MoE layer & $N=K+n^* \times M$ \\
$K$ & Number of base experts duplicated from the retained layer's native MLP & Positive integer \\
$M$ & Number of supplementary experts duplicated per redundant layer's MLP & Positive integer \\
$k$ & Number of Top-K activated experts in MoE routing & Positive integer, default $k=1$ \\
$\delta$ & Inter-layer cosine similarity threshold & Value in $(0,1)$ \\
$\epsilon$ & Inter-layer feature norm relative difference threshold & Value in $(0,1)$ \\
$\tau$ & Routing gating temperature coefficient & Value $\tau>0$, fixed to $\tau=1.0$ for inference \\
$\alpha$ & Load balancing auxiliary loss weight & Positive real number, our default value $\alpha=0.001$ \\

$m$ & Number of pruning depth for sensitivity analysis & Positive real number, $m \in \{1,2,\dots,L\}$ \\
$LaT$ & Roofline theoretical latency & Positive real number \\

\bottomrule
\end{tabular}
\end{table} 

\newpage

\section{Algorithm Details}
\label{app:algorithm}

\begin{algorithm}[htbp]
\caption{Global scoring-based redundant block search}
\label{alg:global_search}
\begin{algorithmic}[1]
\Require Output similarity matrix $S_{out} \in \mathbb{R}^{L \times L}$, MLP similarity matrix $S_{mlp} \in \mathbb{R}^{L \times L}$, norm relative error matrix $\Delta_{norm} \in \mathbb{R}^{L \times L}$, thresholds $\delta, \epsilon$, target continuous block sizes $\mathcal{T}$ (e.g., $\{1, 2, 3\}$), penalty coefficient $\lambda$
\Ensure Kept layers $\mathcal{L}_{keep}$, pruned layers $\mathcal{L}_{prune}$, fused block mappings $\mathcal{B}$

\State Initialize candidate block list $C \leftarrow \emptyset$ \Comment{Phase 1: Exhaustive generation of valid blocks}
\For{$l = 1$ \textbf{to} $L$}
    \For{$n \in \mathcal{T}$}
        \If{$l + n \le L$ \textbf{and} \textsc{IsValidBlock}($S_{out}, S_{mlp}, \Delta_{norm}, l, n, \delta, \epsilon$)}
            \State $\mathit{Score} \leftarrow \frac{1}{n} \sum_{k=1}^{n} \left[ \frac{S_{out}[l, l+k] + S_{mlp}[l, l+k]}{2} - \lambda \Delta_{norm}[l, l+k] \right]$
            \State Add tuple $(Score, l, n)$ \textbf{to} $C$
        \EndIf
    \EndFor
\EndFor

\Statex \hfill $\triangleright$ Phase 2: Global non-overlapping optimal matching
\State Sort $C$ in descending order by $\mathit{Score}$
\State Initialize $Occupied \leftarrow [\textbf{False}] \times L$
\State Initialize $\mathcal{B} \leftarrow \emptyset$, $\mathcal{L}_{prune} \leftarrow \emptyset$

\ForAll{$(Score, l, n) \in C$} 
    \If{$\forall i \in [l, l+n], Occupied[i] == \textbf{False}$}
        \State Add block mapping $(l, \{l+1, \dots, l+n\})$ \textbf{to} $\mathcal{B}$
        \State $\mathcal{L}_{prune} \leftarrow \mathcal{L}_{prune} \cup \{l+1, \dots, l+n\}$
        \State $Occupied[l \dots l+n] \leftarrow \textbf{True}$
    \EndIf
\EndFor

\Statex \hfill $\triangleright$ Phase 3: Extract retained layers
\State $\mathcal{L}_{keep} \leftarrow \{ i \mid i \in \{1, \dots, L\} \land i \notin \mathcal{L}_{prune} \}$
\State \Return $\mathcal{L}_{keep}, \mathcal{L}_{prune}, \mathcal{B}$
\end{algorithmic}
\end{algorithm}

\section{Detailed Quantitative Evaluation of Topological Depth}
\label{app:reward_details}

Table~\ref{tab:reward_details} presents the detailed physical latency, average benchmark score, and the calculated hardware-aware reward for each candidate topological depth. The reward is computed using $w \approx -0.15$ and $L_{base}=195.87 ms$, with the 19-layer configuration achieving the maximum overall value.

\begin{table}[htbp]
    \centering
    \caption{Detailed capacity-efficiency metrics across varying retained depths.}
    \label{tab:reward_details}
    \begin{tabular}{cccc}
        \toprule
        \textbf{Retained Layers ($L$)} & \textbf{Latency $LaT_i$ (ms)} & \textbf{Average Score $S_i$} & \textbf{Calculated Reward} \\
        \midrule
        13 & 135.78 & 29.85 & 31.54 \\
        15 & 148.84 & 39.06 & 40.7 \\
        17 & 161.89 & 42.76 & 44 \\
        \textbf{19} & 174.95 & 47.31 & \textbf{48.12} \\
        21 & 188.00 & 47.50 & 47.79 \\
        23 & 201.05 & 47.93 & 47.74 \\
        \bottomrule
    \end{tabular}
\end{table}

\newpage 
\section{Theoretical latency modeling via Roofline framework}
\label{app:roofline}

In edge-device deployment, empirical end-to-end latency is often obfuscated by sub-optimal low-level kernel implementations (e.g., unoptimized MoE routing or unaligned memory access in inference engines), which mask the true architectural efficiency. To rigorously and objectively compare our co-designed \textbf{Dense2MoE} against the dense \textbf{Seed Model}, we establish a first-principles theoretical latency model grounded in the hardware Roofline framework.

\textbf{Hardware profile and Roofline principle.} Under the Roofline model, the execution time $\mathcal{T}$ of any computational kernel is bottlenecked by either the hardware's peak compute throughput $\pi_H$ (FLOPS) or its sustained memory bandwidth $\beta_H$ (Bytes/s). For our target NVIDIA Jetson Thor-U platform, we adopt the official hardware limits: $\pi_H = 350$~TFLOPS (Tensor Core) and $\beta_H = 273$~GB/s. The physical latency lower bound is formulated as:
\begin{equation}
\mathcal{T} = \max\left(\frac{\mathcal{F}}{\pi_H}, \frac{\mathcal{M}}{\beta_H}\right),
\end{equation}
where $\mathcal{F}$ denotes the total floating-point operations and $\mathcal{M}$ denotes the memory traffic volume. 

\textbf{Phase-wise latency calculation.} The total autoregressive inference latency $\hat{T}_{\theta}$ of an $l$-layer model processing $S_{\text{in}}$ input tokens and generating $S_{\text{out}}$ tokens consists of two distinct phases.

\textbf{Prefill phase (compute-bound).} Processing $S_{\text{in}}$ tokens in parallel heavily utilizes matrix multiplications, rendering it compute-bound ($\frac{\mathcal{F}}{\pi_H} > \frac{\mathcal{M}}{\beta_H}$). The compute coefficient is defined as $\xi_F = 4 + 4/\text{gqa} + 6r$, where $r = K \times r_{\text{single}}$ is the effective FFN expansion ratio ($K$ is the number of active experts per token). The prefill latency is:
\begin{equation}
T_{\text{pre}} = l \cdot \frac{S_{\text{in}} d^2 \xi_F}{\pi_H}.
\end{equation}

\textbf{Decode phase (memory-bound).} Generating tokens step-by-step is strictly memory-bound ($\frac{\mathcal{M}}{\beta_H} > \frac{\mathcal{F}}{\pi_H}$). The latency is dominated by loading static weights and dynamic KV-cache. We define the decode weight traffic coefficient as $\xi_W^{\text{dec}} = 2 + 2/\text{gqa} + 3r$. The decode latency is formulated as:
\begin{equation}
T_{\text{dec}} = l \cdot S_{\text{out}} \cdot \frac{\xi_W^{\text{dec}} d^2 b_w + \frac{2\bar{S}d b_{kv}}{\text{gqa}}}{\beta_H},
\end{equation}
where $b_w = b_{kv} = 2$~Bytes (for FP16/BF16), and $\bar{S} = S_{\text{in}} + (S_{\text{out}} + 1)/2$ is the average context length during generation.

The total theoretical end-to-end latency is the sum of both phases:
\begin{equation}
\hat{T}_{\theta} = T_{\text{pre}} + T_{\text{dec}}.
\end{equation}

\textbf{Architectural equivalence: Why Dense2MoE is strictly faster.} A critical insight from our Roofline formulation is that the total number of experts $E$ does not affect the inference latency. Since Dense2MoE employs a Top-1 routing strategy ($k=1$), its effective expansion ratio $r$ is mathematically identical to the dense Seed Model ($r = r_{\text{single}}$). Consequently, both the compute coefficient $\xi_F$ and the decode traffic coefficient $\xi_W^{\text{dec}}$ are exactly equal for a single Dense2MoE layer and a single Dense layer. The unactivated experts solely consume static memory capacity but incur zero memory bandwidth overhead during forward passes.

By upcycling the 24-layer Dense Seed Model into a 19-layer Dense2MoE, we physically eliminate the attention and active MLP memory traffic of 5 layers. According to the derived equations, this depth reduction ($l: 24 \rightarrow 19$) strictly guarantees a lower physical latency bound ($\hat{T}_{\text{Dense2MoE}} < \hat{T}_{\text{Seed}}$). Therefore, the theoretical latency serves as a fair, hardware-agnostic metric to construct our Pareto frontier, isolating our architectural innovation from the transient inefficiencies of current MoE software kernels.

\newpage
\section{Dataset composition and preprocessing}
\label{app:data}

\begin{table}[htbp]
    \centering
    \small
    \setlength{\tabcolsep}{4pt}
    \setlength{\belowcaptionskip}{8pt}
    \caption{Domain data sources and sampling ratios.}
    \label{tab:data-mixture-app}
    \begin{tabular}{l l c}
        \toprule
        Domain & Dataset & Sampling ratio (\%) \\
        \midrule
        \multirow{3}{*}{Math} & OpenWebMath & 10.0 \\
                              & Arxiv & 12.0 \\
                              & Github & 2 \\
        \midrule
        \multirow{2}{*}{Code} & Code & 16.2 \\
                              & Synthetic Data & 3.8 \\
        \midrule
        \multirow{3}{*}{General knowledge} & Wikipedia & 46.6 \\
                                           & COIG & 6.8 \\
                                           & C4 & 2.6 \\
        \bottomrule
    \end{tabular}
\end{table}

\section{Ablation studies}
\label{app:ablation}

\subsection{Necessity of the Feature Norm Constraint (\texorpdfstring{$\epsilon$}{epsilon})}
\label{app:ablation_norm}

To empirically validate the necessity of the feature norm constraint ($\epsilon$) introduced in Section~\ref{sec:similarity_and_tradeoff}, we conducted a component-wise ablation during the initial phase of continual pre-training. We compared our standard Dense2MoE layer selection strategy (which enforces both $\delta$ and $\epsilon$) against a baseline relying solely on the cosine similarity threshold ($\delta$).

When the norm constraint is omitted, the selection algorithm occasionally fuses layers that possess highly aligned directional features but vastly different activation magnitudes. During dynamic MoE routing, switching between these unconstrained experts introduces abrupt scale shifts into the residual stream. Consequently, in the $\delta$-only setting, we observed pronounced gradient instability and a significantly higher initial training loss (approximately 15\% higher spike) during the first 1,000 steps of continual pre-training. 

In contrast, enforcing our default norm constraint ($\epsilon$) mathematically bounds these magnitude variations, ensuring that all upcycled experts operate within a mutually compatible numerical range. This mechanism effectively stabilizes the residual stream, enabling smooth and rapid convergence from the very first training step. This confirms that while $\delta$ captures semantic redundancy, $\epsilon$ acts as a non-negotiable physical guardrail for numerical stability.

\subsection{Inter-Layer Similarity Visualization}
\label{app:similarity_viz}

\begin{figure}[htbp]
    \centering
    \small
    \setlength{\tabcolsep}{2pt}
    \begin{tabular}{c *{3}{c}}
        & {Qwen2.5-0.5B} & {Qwen2.5-1.5B} & {Llama2-7B} \\
        \hline
        \rotatebox{90}{\shortstack{{Normalized MLP }\\{input similarity}}} &
        \includegraphics[height=0.1\textheight, keepaspectratio]{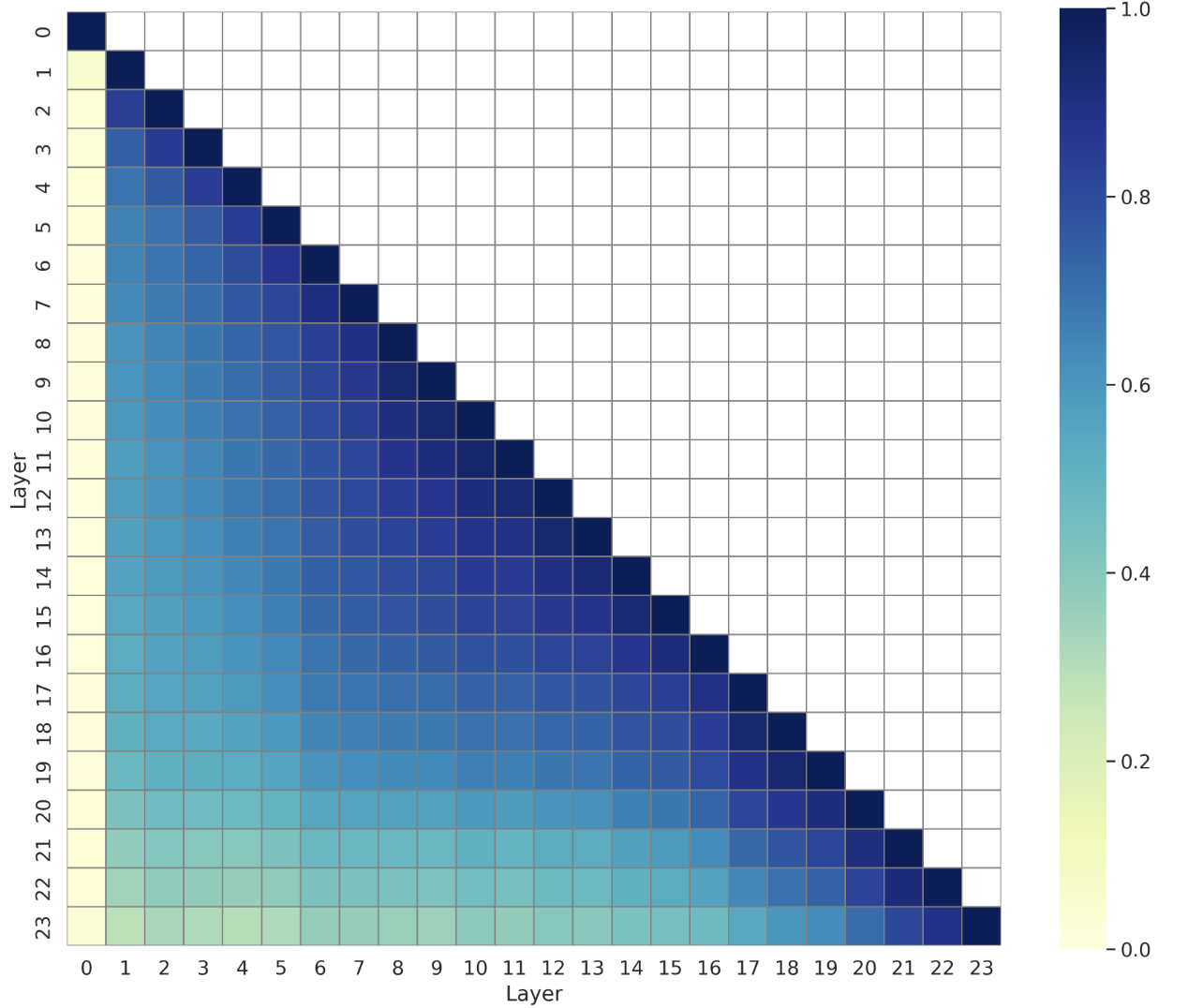} &
        \includegraphics[height=0.1\textheight, keepaspectratio]{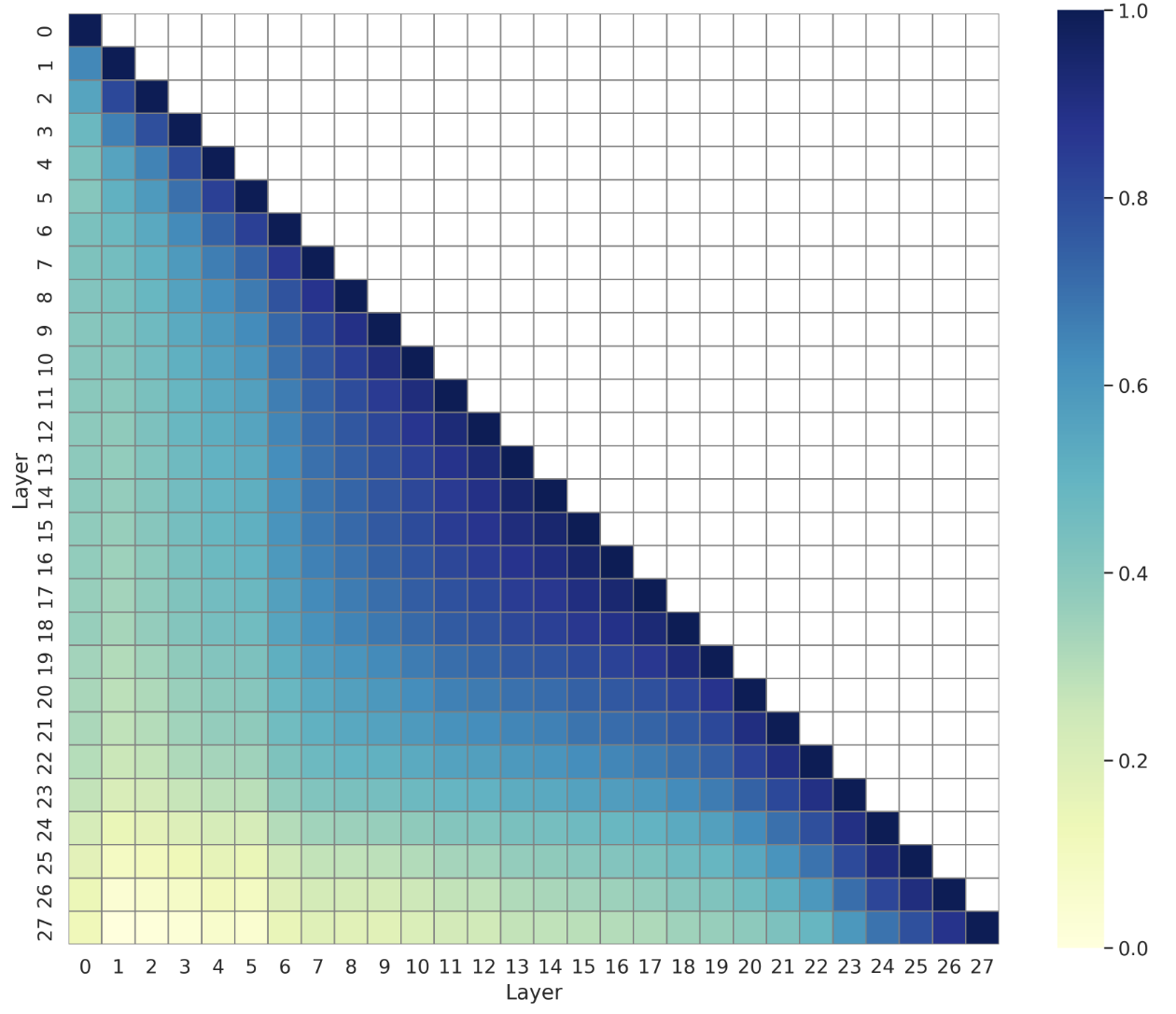} &
        \includegraphics[height=0.1\textheight, keepaspectratio]{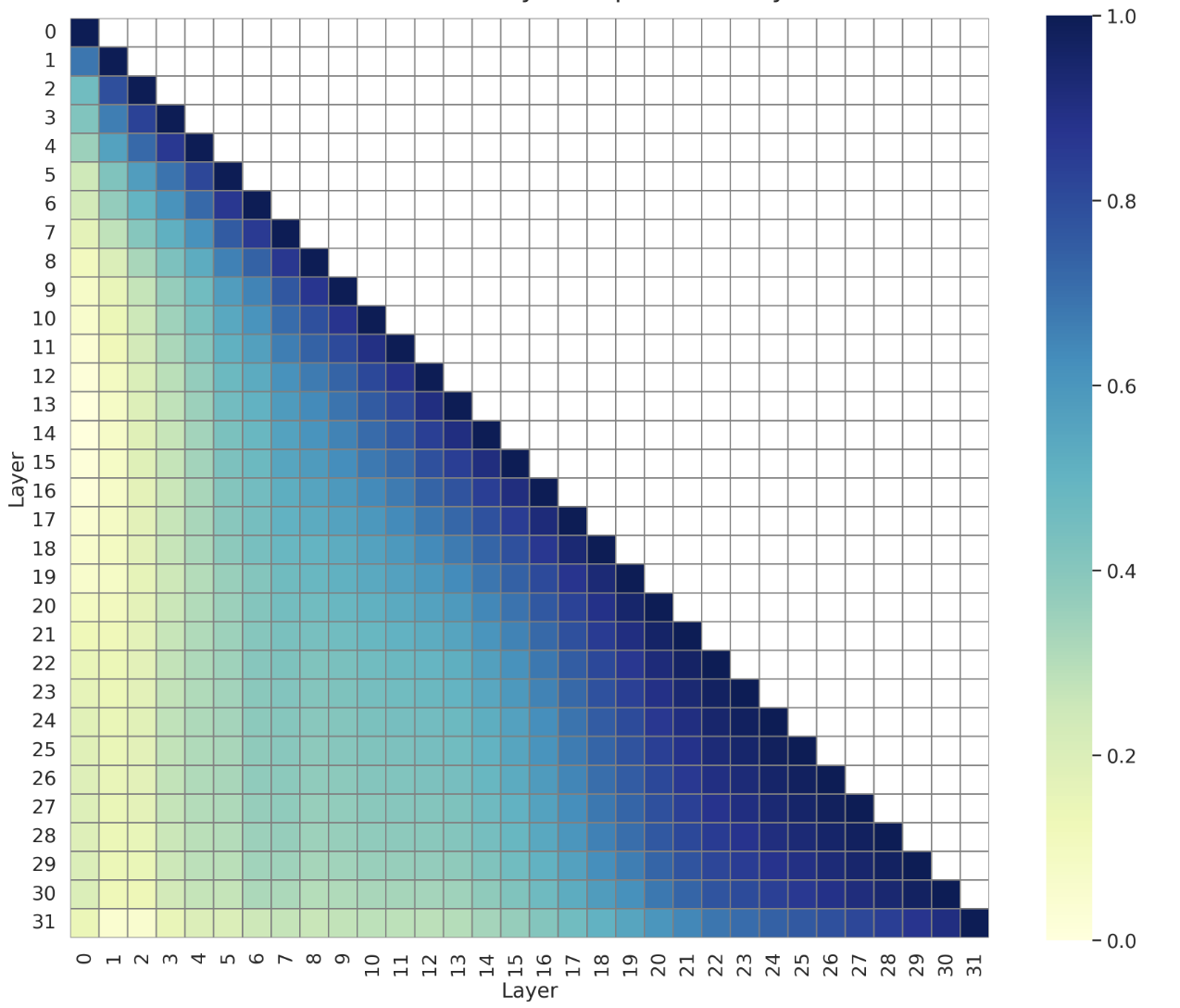} \\
        \hline
        \rotatebox{90}{\shortstack{{Normalized inter-layer }\\{output similarity}}} &
        \includegraphics[height=0.1\textheight, keepaspectratio]{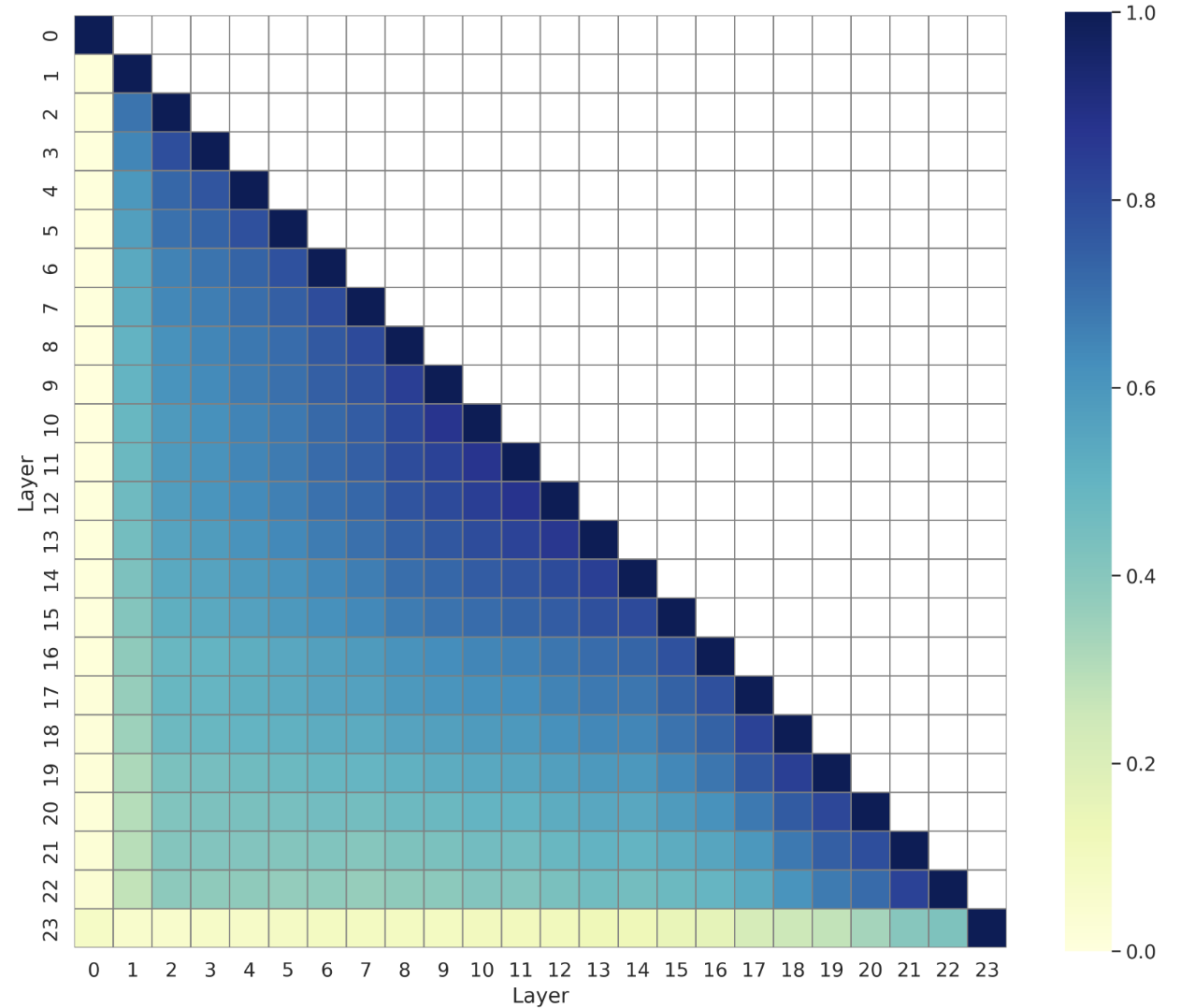} &
        \includegraphics[height=0.1\textheight, keepaspectratio]{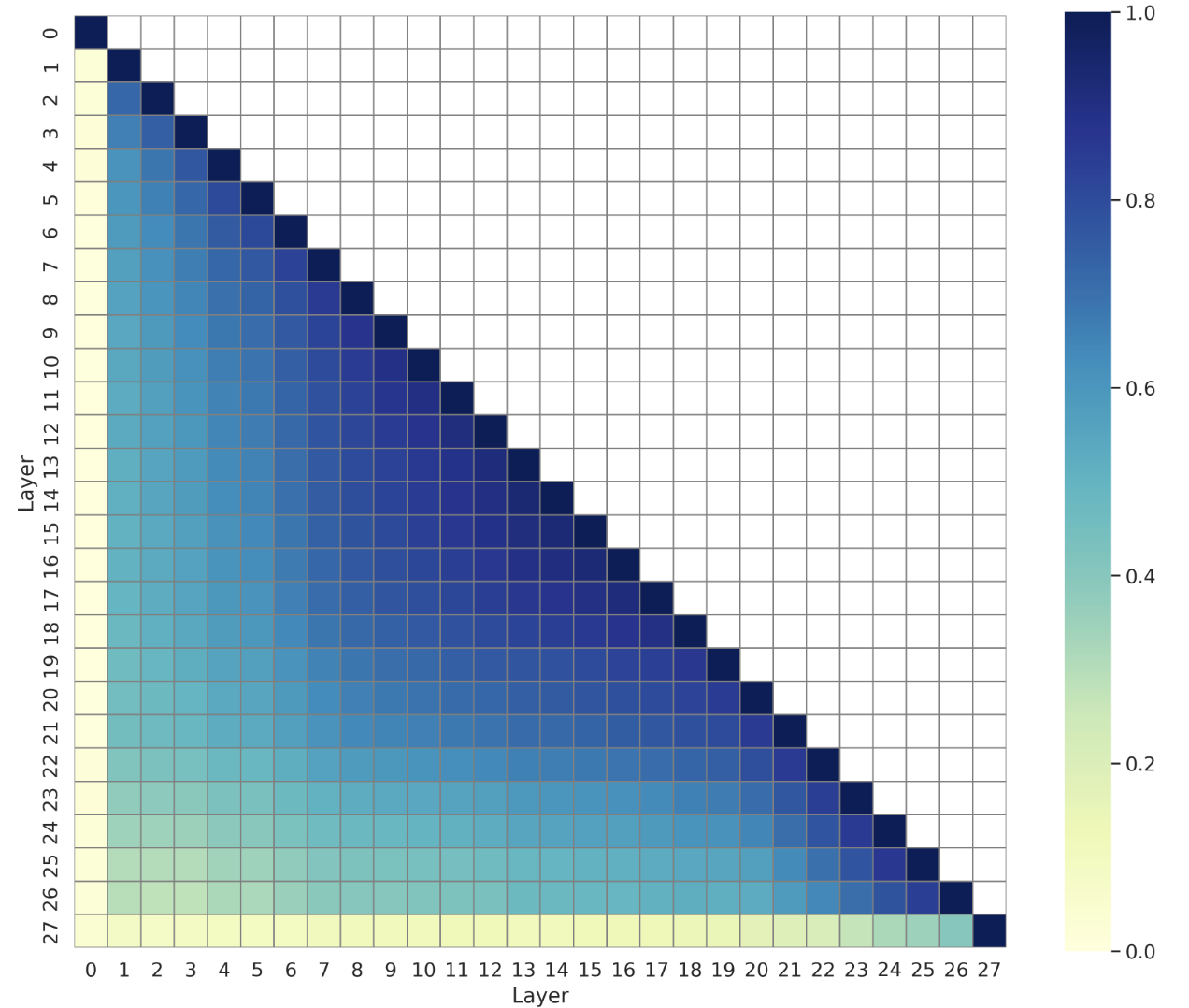} &
        \includegraphics[height=0.1\textheight, keepaspectratio]{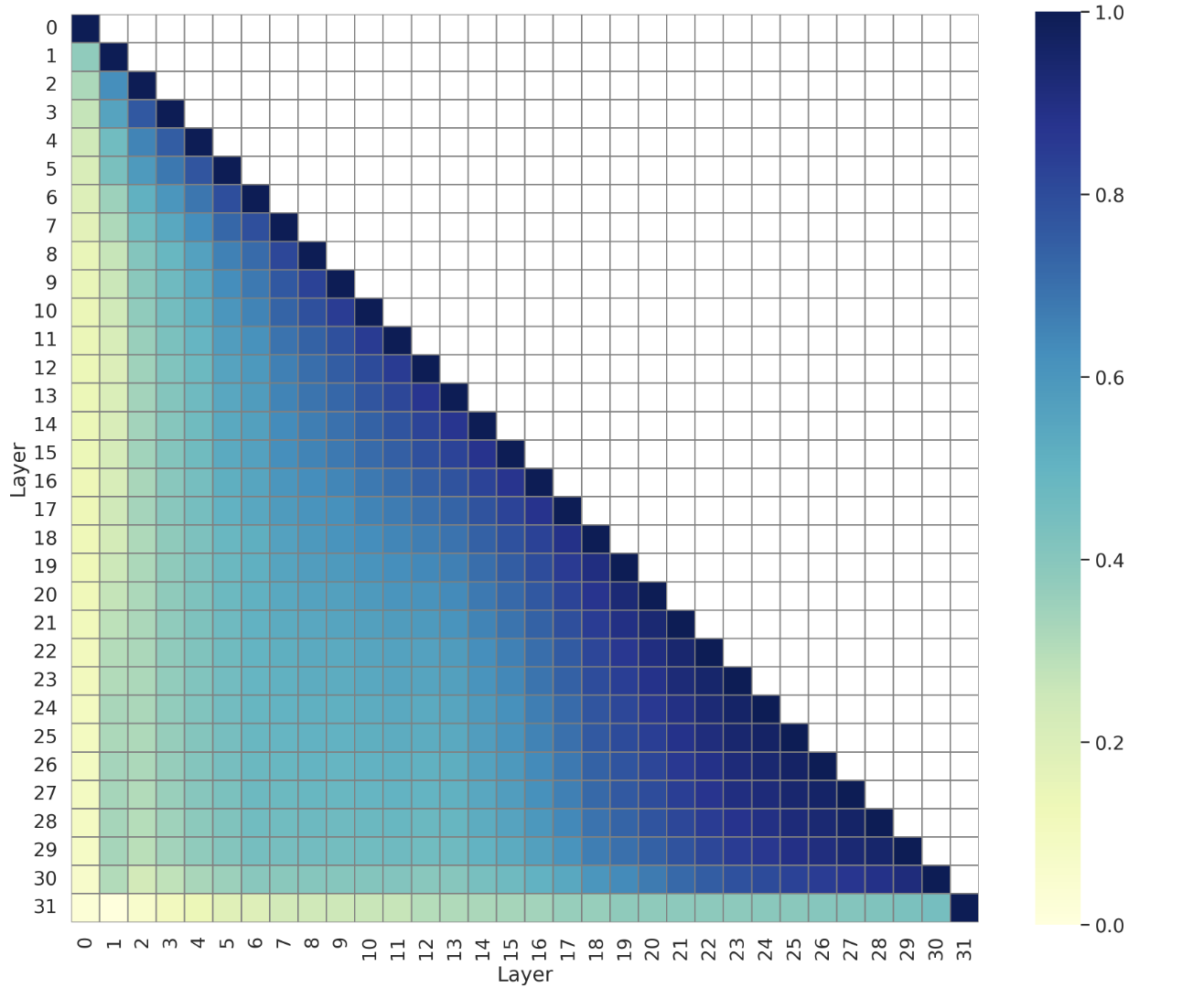} \\
        \hline
    \end{tabular}
    \caption{Normalized similarity comparison across different models.}
    \label{fig3}
\end{figure}

\subsection{Sensitivity to pruning depth thresholds}
\label{app:threshold_sensitivity}

\begin{figure}[htbp]
    \centering
    \includegraphics[width=0.6\textwidth]{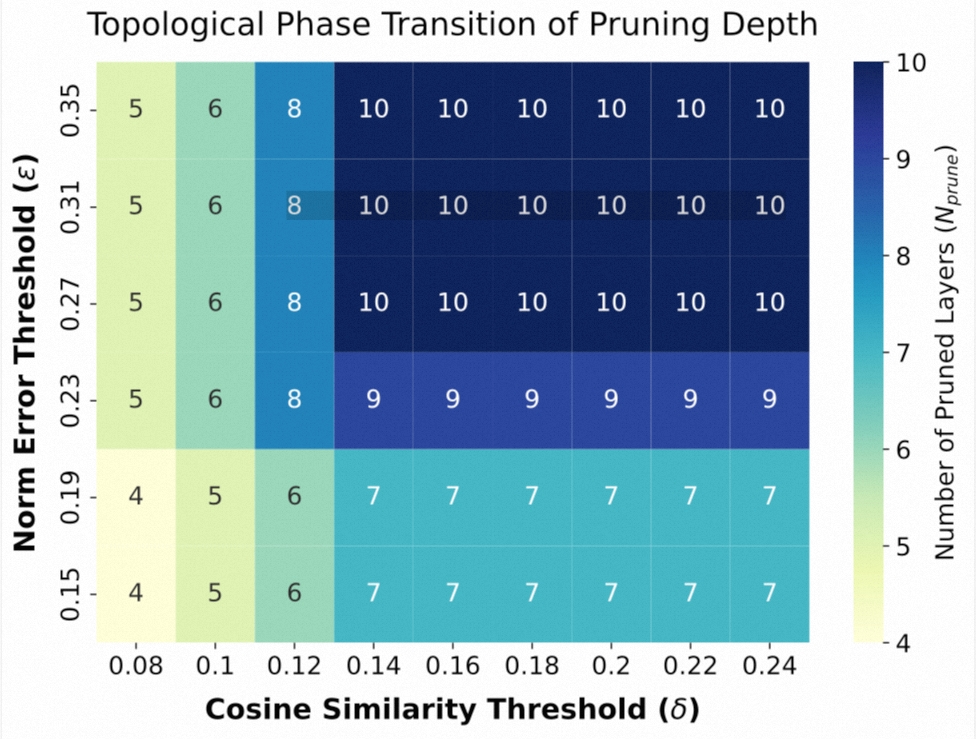}
    \caption{Hyperparameter to pruning depth mapping.}
    \label{fig:heatmap-app} 
\end{figure}

The mapping $m = f(\delta, \epsilon)$ enables systematic control over structural compression. Tighter thresholds preserve more layers, while relaxed thresholds aggressively prune redundant blocks. This discrete phase transition guides our selection of the optimal trade-off point ($m=5$) for edge deployment.

\subsection{Training dynamics and efficiency trade-offs}
\label{app:training_dynamics}

This section provides the detailed visualizations supporting the efficiency-capacity trade-off discussed in Section~\ref{sec:efficiency_tradeoff}. 

\begin{figure}[htbp]
    \centering
    \includegraphics[width=0.6\textwidth]{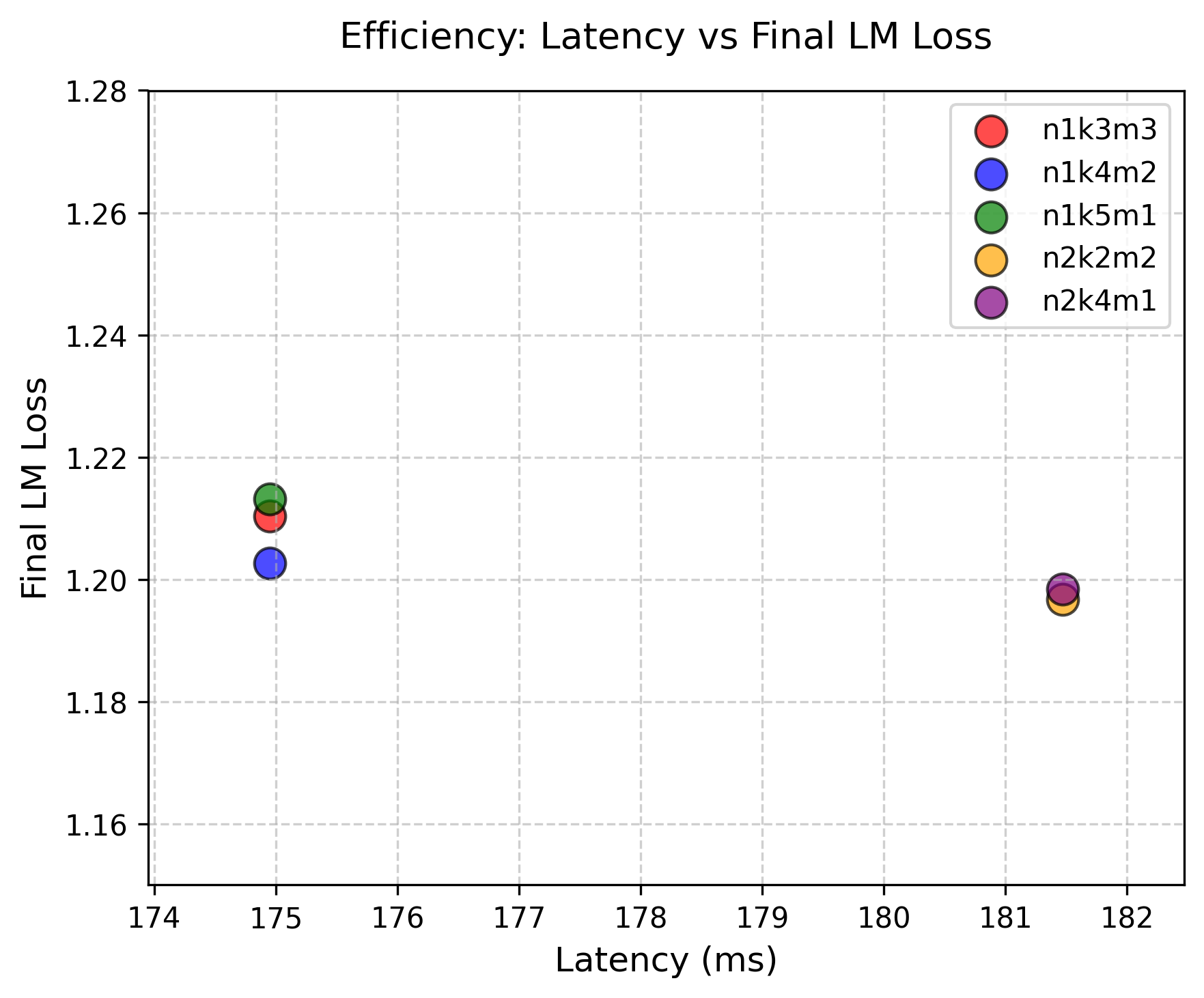}
    \caption{System-level efficiency versus representational capacity.}
    \label{fig:app_efficiency_scatter}
\end{figure}

\begin{figure}[htbp]
    \centering
    \includegraphics[width=0.65\linewidth]{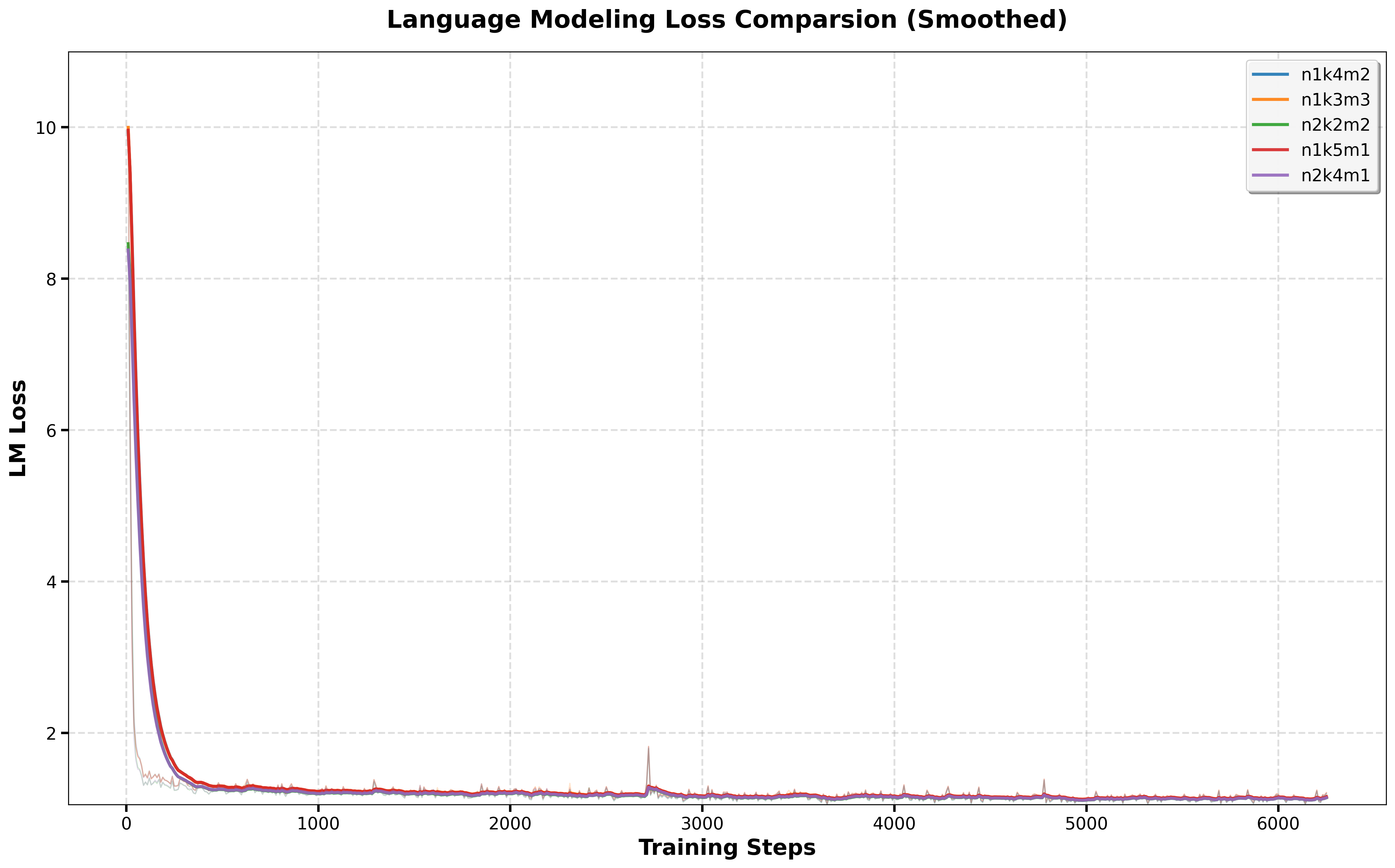}
    \caption{Comparison of causal LM loss across different expert allocations during the continuous pre-training phase.}
    \label{fig:app_lm_loss}
\end{figure}

\begin{figure}[htbp]
    \centering
    \includegraphics[width=0.65\linewidth]{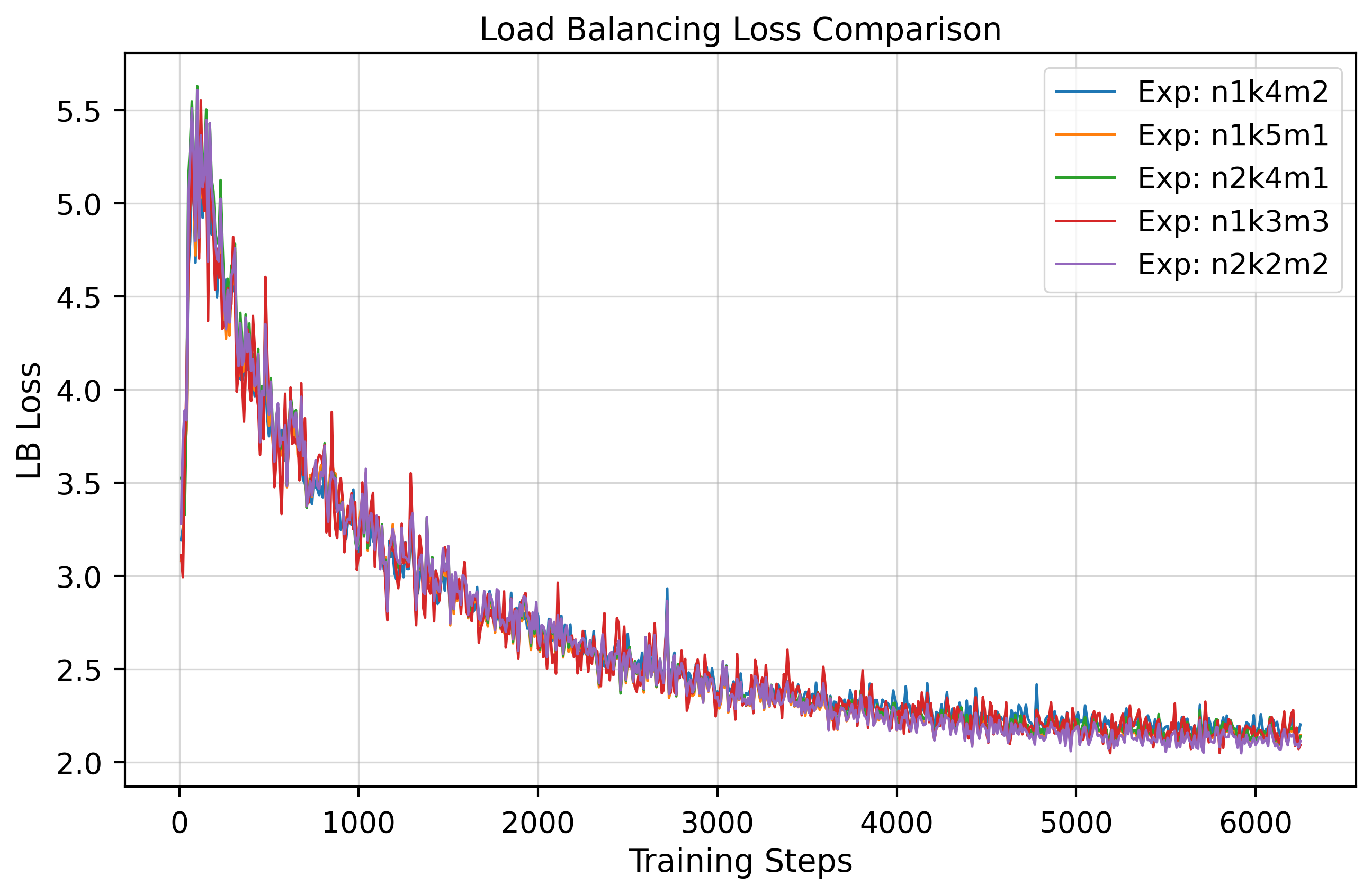}
    \caption{Comparison of load balancing loss across different expert allocations.}
    \label{fig:app_lb_loss}
\end{figure}

Figure~\ref{fig:app_efficiency_scatter} illustrates the system-level efficiency versus representational capacity, highlighting the optimal positioning of the \textbf{n1k4m2} configuration. Furthermore, Figures~\ref{fig:app_lm_loss} and \ref{fig:app_lb_loss} present the smoothed Language Modeling (LM) loss and Load Balancing (LB) loss trajectories over 6,000 continuous pre-training steps. These curves empirically validate that our extreme asymmetric expert allocation incurs no convergence penalty and effectively prevents routing collapse.

\subsection{Static Parameter Memory Derivation}
\label{app:memory_formula}

The static parameter memory (BF16) reported in Figure~\ref{fig:expert_scaling} is computed analytically from the Qwen2.5-0.5B architecture pruned to $L=19$ MoE layers:

\begin{equation}
M = \underbrace{Vd}_{\text{embed}} +
    L\!\Bigl(
      \underbrace{(n_h + 2n_{\mathrm{kv}})d_h d + d^2}_{\text{attention }(Q\!+\!K\!+\!V\!+\!O)}
    + \underbrace{2d_h}_{\text{QK-norm}}
    + \underbrace{2d}_{\text{layer norms}}
    + \underbrace{Nd}_{\text{router gate}}
    + \underbrace{3Nd\,d_{\mathrm{ffn}}}_{\text{expert MLPs}}
    \Bigr)
    + d
\label{eq:mem}
\end{equation}

\noindent where $V{=}151936$, $d{=}896$, $n_h{=}14$, $n_{\mathrm{kv}}{=}2$, $d_h{=}64$, $d_{\mathrm{ffn}}{=}4864$, and $N$ is the number of experts. Model size in GB equals $M \times 2 / 10^9$ (BF16, 2~bytes per parameter). The formula counts static weights only; runtime activations and KV cache are excluded as they depend on batch size and sequence length.

\subsection{Winner-Takes-All Routing Analysis}
\label{app:wta}

\begin{figure}[htbp]
    \centering
    \includegraphics[width=\linewidth]{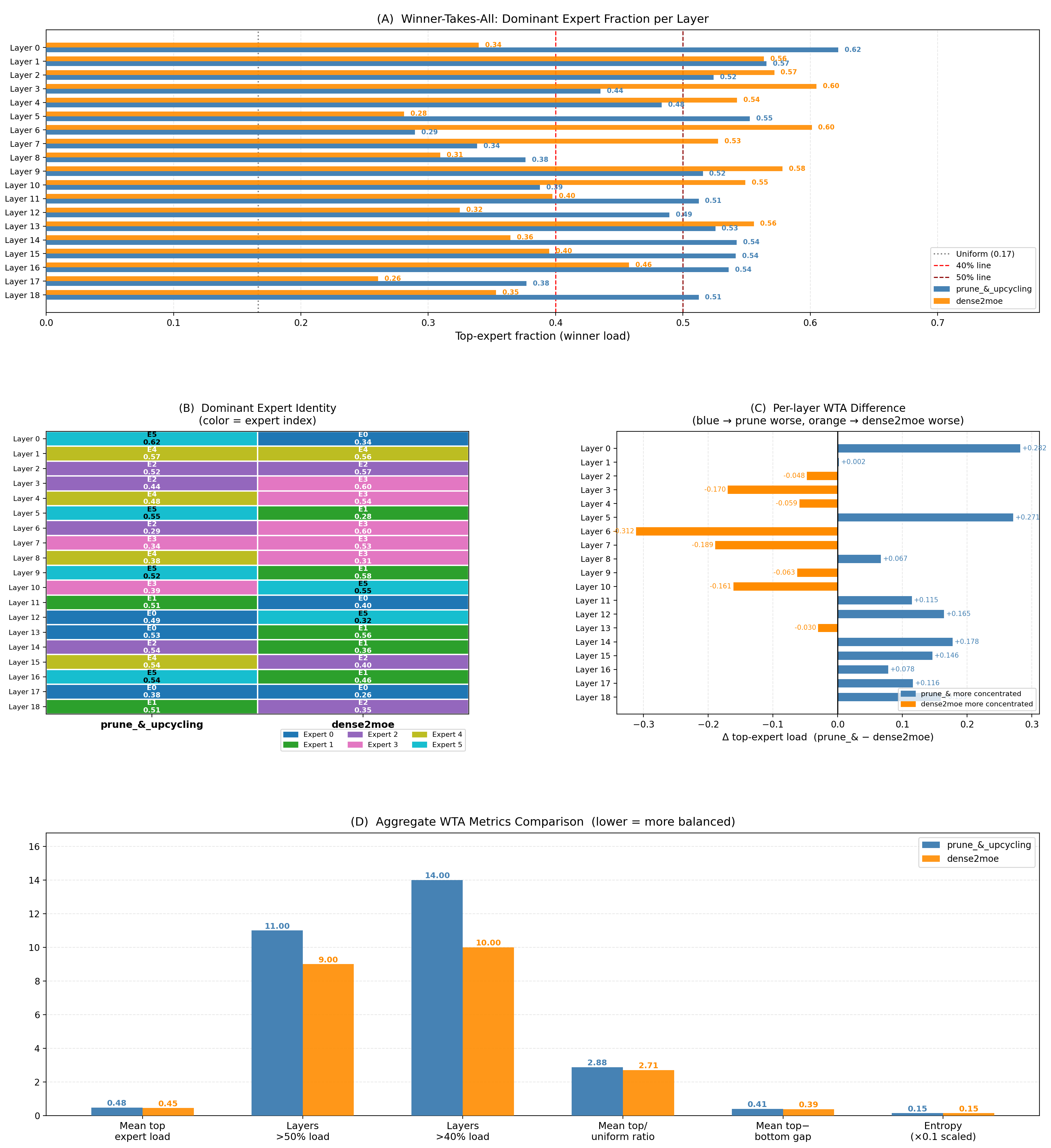}
    \caption{
        \textbf{Winner-Takes-All analysis: Prune+Upcycling vs.\ Dense2MoE (LF-UC).}
        \textbf{(A)}~Dominant-expert load fraction per layer (dashed: 40\%/50\% thresholds).
        \textbf{(B)}~Identity of the winning expert per layer (color = expert index E0--E5).
        \textbf{(C)}~Per-layer load difference (Prune+Upcycling $-$ Dense2MoE); blue: Prune+Upcycling more concentrated; orange: Dense2MoE more concentrated.
        \textbf{(D)}~Aggregate WTA metrics; lower is more balanced except entropy.
    }
    \label{fig:wta_analysis}
\end{figure}

We compare expert routing dynamics between \textbf{Prune+Upcycling} (homogeneous: all experts copied from one MLP) and \textbf{Dense2MoE} (heterogeneous via LF-UC), both with $N{=}6$, top-1 routing, $L{=}19$, evaluated on 1{,}100 held-out samples from five domains (web text, GSM8K, BBH, code, CMMLU).
As shown in Figure~\ref{fig:wta_analysis} and Table~\ref{tab:wta_summary}, Dense2MoE consistently exhibits lower load concentration across all metrics.
Homogeneous initialisation creates a degenerate router from the start: because all experts are identical copies of the same MLP, any early imbalance in token assignment triggers a rich-get-richer feedback loop---the most-visited expert receives the most gradient updates, becomes marginally better, and attracts yet more tokens---causing a single expert to dominate in 11 layers ($>$50\% load) and monopolise routing across consecutive layers (panel B).
LF-UC breaks this feedback loop by sourcing each expert from a \emph{different} pre-trained layer: tokens naturally align better with different layer-specific transformations, so the router finds meaningful load-distribution signal from step one and the auxiliary load-balancing loss~\citep{fedus2022switch} operates on a non-degenerate initial distribution.

\begin{table}[h]
\centering
\small
\caption{Winner-Takes-All routing metrics. Lower is better except entropy ($\uparrow$).}
\label{tab:wta_summary}
\begin{tabular}{lcc}
\toprule
Metric & Prune+Upcycling & Dense2MoE \\
\midrule
Mean top-expert load      & 0.480 & \textbf{0.452} \\
Layers with top $>$ 50\%  & 11    & \textbf{9}     \\
Layers with top $>$ 40\%  & 14    & \textbf{10}    \\
Mean top / uniform ratio  & $2.88\times$ & $\mathbf{2.71\times}$ \\
Mean top$-$bottom gap     & 0.406 & \textbf{0.388} \\
Mean entropy ($\uparrow$) & 1.487 & \textbf{1.489} \\
\bottomrule
\end{tabular}
\end{table}

\end{document}